\documentclass[10pt,twocolumn,letterpaper]{article}

\usepackage{iccv}
\usepackage{times}
\usepackage{epsfig}
\usepackage{graphicx}
\usepackage{amsmath}
\usepackage{amssymb}

\iccvfinalcopy % *** Uncomment this line for the final submission

\usepackage{graphicx}
\usepackage[utf8]{inputenc} % allow utf-8 input
\usepackage[T1]{fontenc}    % use 8-bit T1 fonts
\usepackage{url}            % simple URL typesetting
\usepackage{booktabs}       % professional-quality tables
\usepackage{amsfonts}       % blackboard math symbols
\usepackage{nicefrac}       % compact symbols for 1/2, etc.
\usepackage{microtype}      % microtypography        % colors
\usepackage{epsfig}
\usepackage{float}
\usepackage{multicol}
\usepackage{multirow}
\usepackage{amssymb}
\usepackage{amsmath}
\usepackage{wrapfig}
\usepackage[toc,page]{appendix}
\usepackage{xspace}
\usepackage{color}
\usepackage{colortbl}
\usepackage[dvipsnames, svgnames, x11names]{xcolor}
\usepackage[misc]{ifsym}
\usepackage{makecell}
\usepackage{soul}
\usepackage{bm}
\usepackage{wrapfig}

\definecolor{linkcolor}{RGB}{255,0,0}
\definecolor{urlcolor}{RGB}{255,105,180}
\definecolor{citecolor}{RGB}{66,168,235}
\usepackage[pagebackref=true,breaklinks=true,colorlinks=true,bookmarks=false,linkcolor=linkcolor,urlcolor=urlcolor,citecolor=citecolor]{hyperref}

\usepackage{adjustbox}
\usepackage{pifont}
\usepackage{caption}
\usepackage{array}
\usepackage{enumitem}
\newcolumntype{C}[1]{>{\centering\arraybackslash}p{#1}} % 表格列宽

\def \pzo {\phantom{0}}

\newcommand{\cmark}{\ding{52}\xspace}%
\newcommand{\xmark}{\ding{56}\xspace}%
\newcommand{\xmarkg}{\textcolor{lightgray}{\ding{56}}\xspace}%
\newcommand{\pmark}{\ding{58}\xspace}%

\definecolor{lightgray}{rgb}{0.8, 0.8, 0.8}
\definecolor{lgray}{rgb}{0.66, 0.66, 0.66}
\definecolor{whit_tab}{RGB}{255, 255, 255}
\definecolor{gray_tab}{RGB}{235, 235, 235}
\definecolor{oran_tab}{RGB}{254, 247, 241}
\definecolor{blue_tab}{RGB}{200, 227, 245}
\definecolor{lblu_tab}{RGB}{231, 239, 248}
\newcommand{\red}{\textcolor[RGB]{255, 36, 24}}
\newcommand{\deepred}{\textcolor[RGB]{176, 36, 24}}
\newcommand{\grayer}{\textcolor[RGB]{168, 168, 168}}

\usepackage[ruled,vlined,linesnumbered]{algorithm2e}
\usepackage[capitalize]{cleveref}
\crefname{section}{Sec.}{Secs.}
\Crefname{section}{Section}{Sections}
\Crefname{table}{Table}{Tables}
\crefname{table}{Tab.}{Tabs.}

\newlength\savewidth\newcommand\shline{\noalign{\global\savewidth\arrayrulewidth\global\arrayrulewidth 1pt}\hline\noalign{\global\arrayrulewidth\savewidth}}
\newcommand{\tablestyle}[2]{\setlength{\tabcolsep}{#1}\renewcommand{\arraystretch}{#2}\centering\footnotesize}
\renewcommand{\paragraph}[1]{\vspace{1.25mm}\noindent\textbf{#1}}

\def\iccvPaperID{6684} % *** Enter the ICCV Paper ID here

\ificcvfinal\fi

\begin{document}

%%%%%%%%% TITLE
\title{Rethinking Mobile Block for Efficient Attention-based Models}

\author{Jiangning Zhang\textsuperscript{1,2} 
\quad Xiangtai Li\textsuperscript{3} 
\quad Jian Li\textsuperscript{1} 
\quad Liang Liu\textsuperscript{1} 
\quad Zhucun Xue\textsuperscript{4}  \\
\quad Boshen Zhang\textsuperscript{1} 
\quad Zhengkai Jiang\textsuperscript{1} 
\quad Tianxin Huang\textsuperscript{2} 
\quad Yabiao Wang\textsuperscript{1}\thanks{Corresponding authors.}
\quad Chengjie Wang\textsuperscript{1}\footnotemark[1] \\
\normalsize \textsuperscript{1}{Youtu Lab, Tencent} \quad \textsuperscript{2}{Zhejiang University} \quad \textsuperscript{3}{Peking University} \quad \textsuperscript{4}{Wuhan University} \\
{\tt\small Code: \url{https://github.com/zhangzjn/EMO}}
}

% \thanks{Equal corresponding authors.}
% \footnotemark[1]

% \twocolumn[{%
% \renewcommand\twocolumn[1][]{#1}
% \maketitle
% \begin{center}
% \centering
% \captionsetup{type=figure}
%     \includegraphics[width=1.0\linewidth]{figs/meta_mobile_block.pdf}
%     \caption[figure]{\textbf{Left}: Abstracted unified \deepred{\textbf{\emph{Meta-Mobile Block}}} from \emph{Multi-Head Self-Attention} and \emph{Feed-Forward Network} in Transformer as well as efficient \emph{Inverted Residual Block} in MobileNet-v2. This inductive block can be deduced into specific modules using different \emph{expansion ratio} \deepred{\emph{\bm{$\lambda$}}} and \emph{efficient operator} \deepred{\bm{$\mathcal{F}$}}. Absorbing the experience of light-weight CNN and Transformer, an efficient but effective \deepred{EMO} is designed based on deduced iRMB (\cf Sec~\ref{section:irmb}). \textbf{Right}: \emph{Performance} \vs \emph{FLOPs} comparisons with SoTA Transformer-based methods.}
%     \label{fig:meta_mobile_block}
% \end{center}
% }]

\maketitle
\ificcvfinal\thispagestyle{empty}\fi

% with MetaFormer
% Method conclusion
% 

% \maketitle
% \blfootnote{Jiangning Zhang (186368@zju.edu.cn); Yabiao Wang (caseywang@tencent.com); Chengjie Wang (jasoncjwang@tencent.com)}
% \blfootnote{$\dagger$ Equal corresponding authors.}
\begin{abstract} \label{section:abs}
  This paper focuses on developing modern, efficient, lightweight models for dense predictions while trading off parameters, FLOPs, and performance. 
  Inverted Residual Block (IRB) serves as the infrastructure for lightweight CNNs, but no counterpart has been recognized by attention-based studies. 
  This work rethinks lightweight infrastructure from efficient IRB and effective components of Transformer from a unified perspective, extending CNN-based IRB to attention-based models and abstracting a one-residual Meta Mobile Block (MMB) for lightweight model design. 
  Following simple but effective design criterion, we deduce a modern \textbf{I}nverted \textbf{R}esidual \textbf{M}obile \textbf{B}lock (iRMB) and build a ResNet-like Efficient MOdel (EMO) with only iRMB for down-stream tasks. 
  Extensive experiments on ImageNet-1K, COCO2017, and ADE20K benchmarks demonstrate the superiority of our EMO over state-of-the-art methods, \eg, EMO-1M/2M/5M achieve 71.5, 75.1, and 78.4 Top-1 that surpass equal-order CNN-/Attention-based models, while trading-off the parameter, efficiency, and accuracy well: running 2.8-4.0$\times\uparrow$ faster than EdgeNeXt on iPhone14. 
  % The code is available at \url{https://github.com/zhangzjn/EMO}.
  
\end{abstract}

\section{Introduction} \label{section:intro}

With a recent increasing demand for storage/computing restricted applications, mobile models with \emph{fewer parameters} and \emph{low FLOPs} have attracted significant attention from developers and researchers. 
The earliest attempt to design an efficient model dates back to the Inceptionv3~\cite{inceptionv3} era, which uses asymmetric convolutions to replace standard convolution. 
Then, MobileNet~\cite{mnetv1} proposes \emph{depth-wise separable convolution} to significantly decrease the amount of computation and parameters, which is viewed as a fundamental CNN-based component for subsequent works~\cite{shufflenetv1,shufflenetv2,espnetv2,ghostnet}. 
Remarkably, MobileNetv2~\cite{mnetv2} proposes an efficient \emph{Inverted Residual Block} (IRB) based on \emph{Depth-Wise Convolution} (DW-Conv) that is recognized as the infrastructure of efficient models~\cite{efficientnet} until now. 
Inevitably, limited by the natural induction bias of static CNN, the \textit{accuracy} of CNN-pure models still maintains a low level of accuracy that needs further improvements.
In summary, one extreme core is to \emph{advance a stronger fundamental block going beyond IRB}. 

\begin{figure}[tp]
    \centering
    \includegraphics[width=0.93\linewidth]{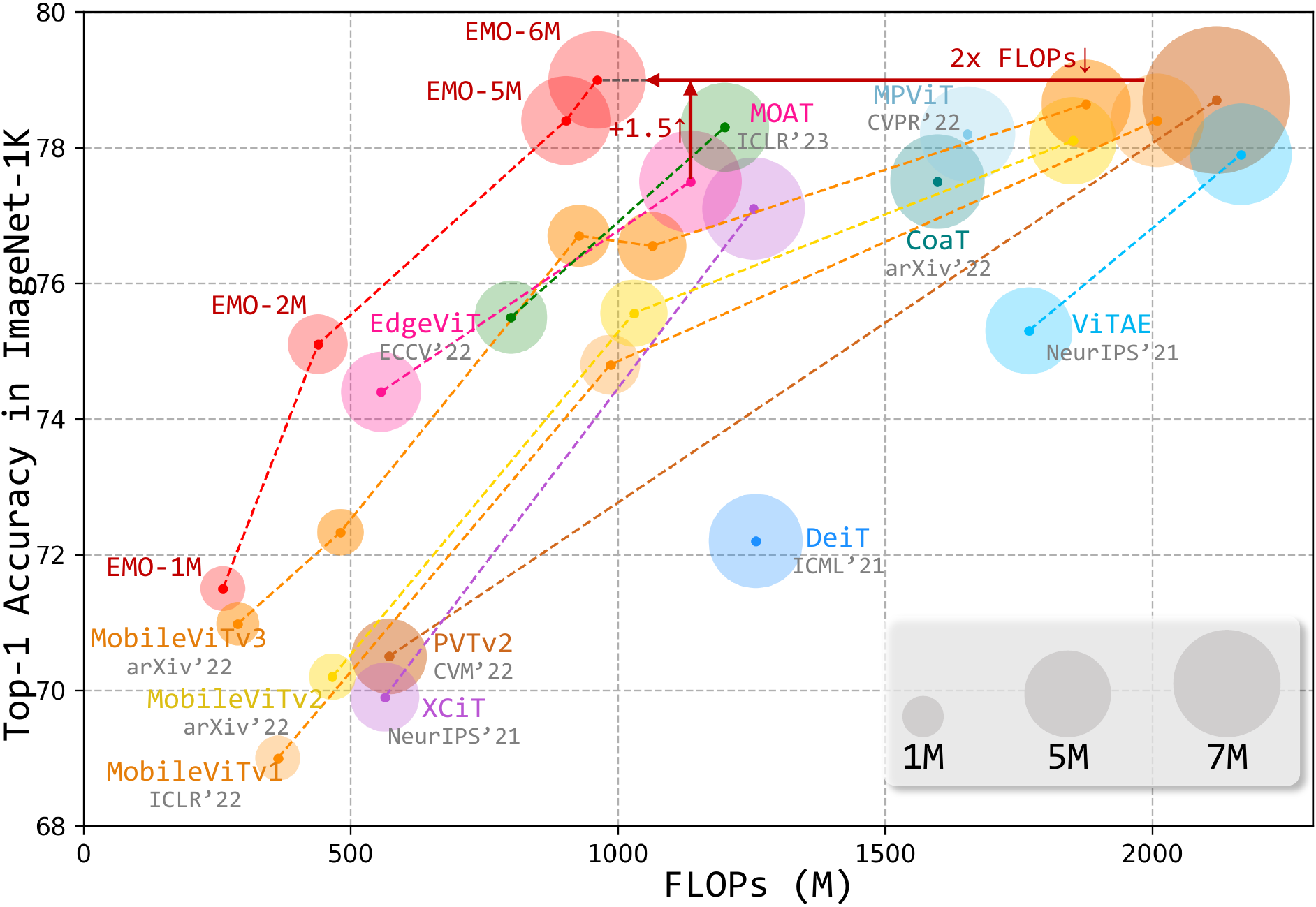}
    \caption{\emph{Performance} \vs \emph{FLOPs} with concurrent methods.}
    \label{fig:sota}
    % \vspace{-2.0em}
\end{figure}
On the other hand, stared from vision transformer (ViTs)~\cite{vit}, many follow-ups~\cite{deit,pvtv1,pvtv2,swinv1,swinv2,zhang2021analogous,zhang2022eatformer,li2023transformer} have achieved significant improvements over CNN.
This is due to its ability to model dynamically and learn from the extensive dataset, and how to migrate this capability to lightweight CNN is worth our explorations. However, limited by the quadratic amount of computations for Multi-Head Self-Attention (MHSA), the attention-based model requires massive resource consumption, especially when the channel and resolution of the feature map are large. 
Some works attempt to tackle the above problems by designing variants with linear complexity~\cite{reformer,performer}, decreasing the spatial resolution of features~\cite{cvt,pvtv1,nextvit}, rearranging channel~\cite{delight}, using local window attention~\cite{swinv1} \etc. However, these methods still cannot be deployed on devices. 

Recently, researchers have aimed to design efficient hybrid models with lightweight CNNs, and they obtain better performances than CNN-based models with trading off accuracy, parameters, and FLOPs. 
However, current methods introduce complex structures~\cite{mvitv1,mvitv2,mvitv3,mobileformer,edgenext} or multiple hybrid modules~\cite{edgenext,edgevit}, which is very detrimental to optimize for applications. 
So far, little work has been done to explore attention-based counterparts as IRB, and this inspires us to think: \emph{Can we build a lightweight IRB-like infrastructure for attention-based models with only basic operators?}

\begin{figure*}[htp]
    \centering
    \includegraphics[width=1.0\linewidth]{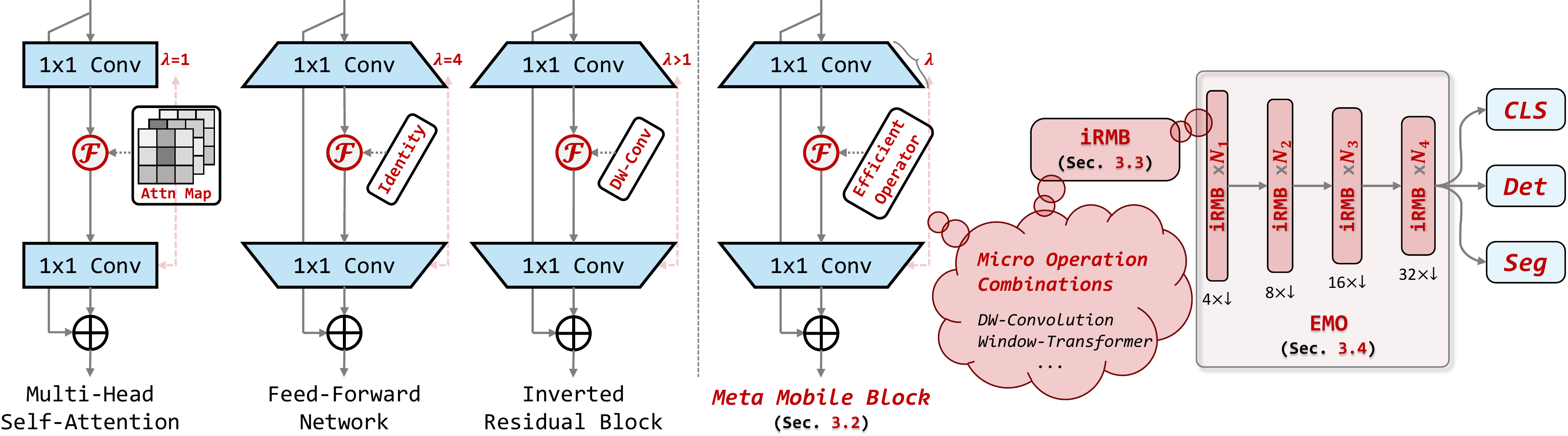}
    \caption{\textbf{Left}: Abstracted unified \deepred{\textbf{\emph{Meta-Mobile Block}}} from \emph{Multi-Head Self-Attention} / \emph{Feed-Forward Network}~\cite{transformer} and \emph{Inverted Residual Block}~\cite{mnetv2} (\cf Sec~\ref{section:mmb}). The inductive block can be deduced into specific modules using different \emph{expansion ratio} \deepred{\emph{\bm{$\lambda$}}} and \emph{efficient operator} \deepred{\bm{$\mathcal{F}$}}. \textbf{Right}: ResNet-like \textit{EMO} composed of only deduced \emph{iRMB} (\cf Sec~\ref{section:irmb}).}
    \label{fig:emo}
    % \vspace{-1.5em}
\end{figure*}

Based on the above motivation, we rethink efficient \emph{Inverted Residual Block} in MobileNetv2~\cite{mnetv2} and effective MHSA/FFN modules in Transformer~\cite{transformer} from a unified perspective, expecting to integrate both advantages at the infrastructure design level. As shown in Fig.~\ref{fig:emo}-Left, while working to bring one-residual IRB with inductive bias into the attention model, we observe two underlying submodules (\ie, FFN and MHSA) in two-residual Transformer share the similar structure to IRB. Thus, we inductively abstract a one-residual Meta Mobile Block (MMB, \cf, Sec.~\ref{section:mmb}) that takes parametric arguments \emph{expansion ratio} \deepred{\emph{\bm{$\lambda$}}} and \emph{efficient operator} \deepred{\bm{$\mathcal{F}$}} to instantiate different modules, \ie, IRB, MHSA, and FFN. We argue that \textit{MMB can reveal the consistent essence expression of the above three modules, and it can be regarded as an improved lightweight concentrated aggregate of Transformer}. Furthermore, a simple yet effective \emph{Inverted Residual Mobile Block} (iRMB) is deduced that only contains fundamental Depth-Wise Convolution and our improved EW-MHSA (\cf, Sec.~\ref{section:irmb}) and we build a ResNet-like 4-phase Efficient MOdel (EMO) with only iRMBs (\cf, Sec.~\ref{section:emo}). 
Surprisingly, our method performs better over the SoTA lightweight attention-based models even without complex structures, as shown in Fig.~\ref{fig:sota}. 
In summary, this work follows simple design criteria while gradually producing an efficient attention-based lightweight model. 

Our contributions are four folds: 
\noindent\textbf{1)} We extend CNN-based IRB to the two-residual transformer and abstract a one-residual \textit{Meta Mobile Block} (MMB) for lightweight model design. This meta paradigm could describe the current efficient modules and is expected to have the guiding significance in concreting novel efficient modules. 
\noindent\textbf{2)} Based on inductive MMB, we deduce a simple yet effective modern \emph{Inverted Residual Mobile Block} (iRMB) and build a ResNet-like Efficient MOdel (EMO) with only iRMB for down-stream applications. In detail, iRMB only consists of naive DW-Conv and the improved EW-MHSA to model short-/long-distance dependency, respectively. 
\noindent\textbf{3)} We provide detailed studies of our method and give some experimental findings on building attention-based lightweight models, hoping our study will inspire the research community to design powerful and efficient models. 
\noindent\textbf{4)} Even without introducing complex structures, our method still achieves very competitive results than concurrent attention-based methods on several benchmarks, \eg, our EMO-1M/2M/5M reach 71.5, 75.1, and 78.4 Top-1 over current SoTA CNN-/Transformer-based models. Besides, EMO-1M/2M/5M armed SSDLite obtain 22.0/25.2/27.9 mAP with only 2.3M/3.3M/6.0M parameters and 0.6G/0.9G/1.8G FLOPs, which exceeds recent MobileViTv2~\cite{mvitv2} by +0.8$\uparrow$/+0.6$\uparrow$/+0.1$\uparrow$ with decreased FLOPs by -33\%$\downarrow$/-50\%$\downarrow$/-62\%$\downarrow$; EMO-1M/2M/5M armed DeepLabv3 obtain 33.5/35.3/37.98 mIoU with only 5.6M/6.9M/10.3M parameters and 2.4G/3.5G/5.8G FLOPs, surpassing MobileViTv2 by +1.6$\uparrow$/+0.6$\uparrow$/+0.8$\uparrow$ with much lower FLOPs.

\section{Methodology: Induction and Deduction} \label{section:method}

\subsection{Criteria for General Efficient Model} \label{section:cri}
When designing efficient visual models for mobile applications, we advocate the following criteria subjectively and empirically that an efficient model should satisfy as much as possible: \textbf{\ding{192} Usability.} Simple implementation that does not use complex operators and is easy to optimize for applications. \textbf{\ding{193} Uniformity.} As few core modules as possible to reduce model complexity and accelerated deployment. \textbf{\ding{194} Effectiveness.} Good performance for classification and dense prediction. \textbf{\ding{195} Efficiency.} Fewer parameters and calculations with accuracy trade-off. We make a summary of current efficient models in Tab.~\ref{table:model_criterion}: \emph{1)} Performance of MobileNet series~\cite{mnetv1,mnetv2,mvitv3} is now seen to be slightly lower, and its parameters are slightly higher than counterparts. \emph{2)} Recent MobileViT series~\cite{mvitv1,mvitv2,mvitv3} achieve notable performances, but they suffer from higher FLOPs and slightly complex modules. \emph{3)} EdgeNeXt~\cite{edgenext} and EdgeViT~\cite{edgevit} obtain pretty results, but their basic blocks also consist of elaborate modules. Comparably, the design principle of our EMO follows the above criteria without introducing complicated operations (\cf, Sec.~\ref{section:emo}), but it still obtains impressive results on multiple vision tasks (\cf, Sec.~\ref{section:exp}).

\begin{table}[htp]
    \centering
    \caption{\textbf{Criterion comparison for current efficient models}. \textbf{\ding{192}}: Usability; \textbf{\ding{193}}: Uniformity; \textbf{\ding{194}}: Effectiveness; \textbf{\ding{195}}: Efficiency. \cmark: Satisfied. \pmark: Partially satisfied. \xmark: Unsatisfied.}
    \label{table:model_criterion}
    \renewcommand{\arraystretch}{1.0}
    \setlength\tabcolsep{6.0pt}
    \resizebox{1.0\linewidth}{!}{
        \begin{tabular}{p{4.6cm}<{\raggedright} p{1.0cm}<{\centering} p{1.0cm}<{\centering} p{1.0cm}<{\centering} p{1.0cm}<{\centering}}
            \toprule[0.17em]
            Method \vs Criterion                            & \ding{192}    & \ding{193}    & \ding{194}    & \ding{195}    \\
            \hline
            MobileNet Series~\cite{mnetv1,mnetv2,mvitv3}    & \cmark        & \cmark        & \pmark        & \pmark        \\
            MobileViT Series~\cite{mvitv1,mvitv2,mvitv3}    & \pmark        & \pmark        & \cmark        & \pmark        \\
            EdgeNeXt~\cite{edgenext}                        & \pmark        & \xmark        & \cmark        & \cmark        \\
            EdgeViT~\cite{edgevit}                          & \cmark        & \pmark        & \cmark        & \pmark        \\
            \hline
            EMO (Ours)                                      & \cmark        & \cmark        & \cmark        & \cmark        \\
            \toprule[0.12em]
        \end{tabular}
    }
    % \vspace{-1.5em}
\end{table}

\subsection{Meta Mobile Block} \label{section:mmb}
\noindent\textbf{Motivation.} 1) Recent Transformer-based works~\cite{focal,swinv1,cswin,inception,gmlp,mlpmixer,resmlp} are dedicated to improving spatial token mixing under the MetaFormer~\cite{metaformer} for high-performance network. CNN-based \textit{Inverted Residual Block}~\cite{mnetv2} (IRB) is recognized as the infrastructure of efficient models~\cite{mnetv2,efficientnet}, but little work has been done to explore attention-based counterpart. This inspires us to build a lightweight IRB-like infrastructure for attention-based models. 2) While working to bring one-residual IRB with inductive bias into the attention model, we stumble upon two underlying sub-modules (\ie, FFN and MHSA) in two-residual Transformer that happen to share a similar structure to IRB. \\
\noindent\textbf{Induction.} 
We rethink Inverted Residual Block in MobileNetv2~\cite{mnetv2} with core MHSA and FFN modules in Transformer~\cite{transformer}, and inductively abstract a general Meta Mobile Block (MMB) in Fig.~\ref{fig:emo}, which takes parametric arguments \emph{expansion ratio} \deepred{\emph{\bm{$\lambda$}}} and \emph{efficient operator} \deepred{\bm{$\mathcal{F}$}} to instantiate different modules. We argue that \textit{the MMB can reveal the consistent essence expression of the above three modules, and MMB can be regarded as an improved lightweight concentrated aggregate of Transformer}. Also, this is the basic motivation for our elegant and easy-to-use EMO, which only contains one deduced iRMB absorbing advantages of lightweight CNN and Transformer. Take image input $\boldsymbol{X} (\in\mathbb{R}^{C \times H \times W})$ as an example, MMB firstly use a expansion $\text{MLP}_{e}$ with output/input ratio equaling \deepred{\emph{\bm{$\lambda$}}} to expand channel dimension:

\vspace{-0.3em}
\begin{equation}
    \begin{aligned}
        \boldsymbol{X}_{e} = \text{MLP}_{e}(\boldsymbol{X}) (\in\mathbb{R}^{\deepred{\emph{\bm{$\lambda$}}}C \times H \times W}) \text{.}
    \end{aligned}
\end{equation}
Then, intermediate operator $\deepred{\bm{\mathcal{F}}}$ enhance image features further, \eg, identity operator, static convolution, dynamic MHSA, \etc. Considering that MMB is suitable for efficient network design, we present $\deepred{\bm{\mathcal{F}}}$ as the concept of \emph{efficient operator}, formulated as: 

\vspace{-0.3em}
\begin{equation}
    \begin{aligned}
        \boldsymbol{X}_{f} = \deepred{\bm{\mathcal{F}}}(\boldsymbol{X}_{e}) (\in\mathbb{R}^{\deepred{\emph{\bm{$\lambda$}}}C \times H \times W}) \text{.}
    \end{aligned}
\end{equation}
Finally, a shrinkage $\text{MLP}_{s}$ with inverted input/output ratio equaling \deepred{\emph{\bm{$\lambda$}}} to shrink channel dimension:
\begin{equation}
    \begin{aligned}
        \boldsymbol{X}_{s} = \text{MLP}_{s}(\boldsymbol{X}_{f}) (\in\mathbb{R}^{C \times H \times W}) \text{,}
    \end{aligned}
\end{equation}
where a residual connection is used to get the final output $\boldsymbol{Y} = \boldsymbol{X} + \boldsymbol{X}_{s} (\in\mathbb{R}^{C \times H \times W})$. Notice that normalization and activation functions are omitted for clarity.

\begin{wrapfigure}{r}{3.8cm}
    \centering
    % \vspace{-1.2em}
    \includegraphics[width=1.0\linewidth]{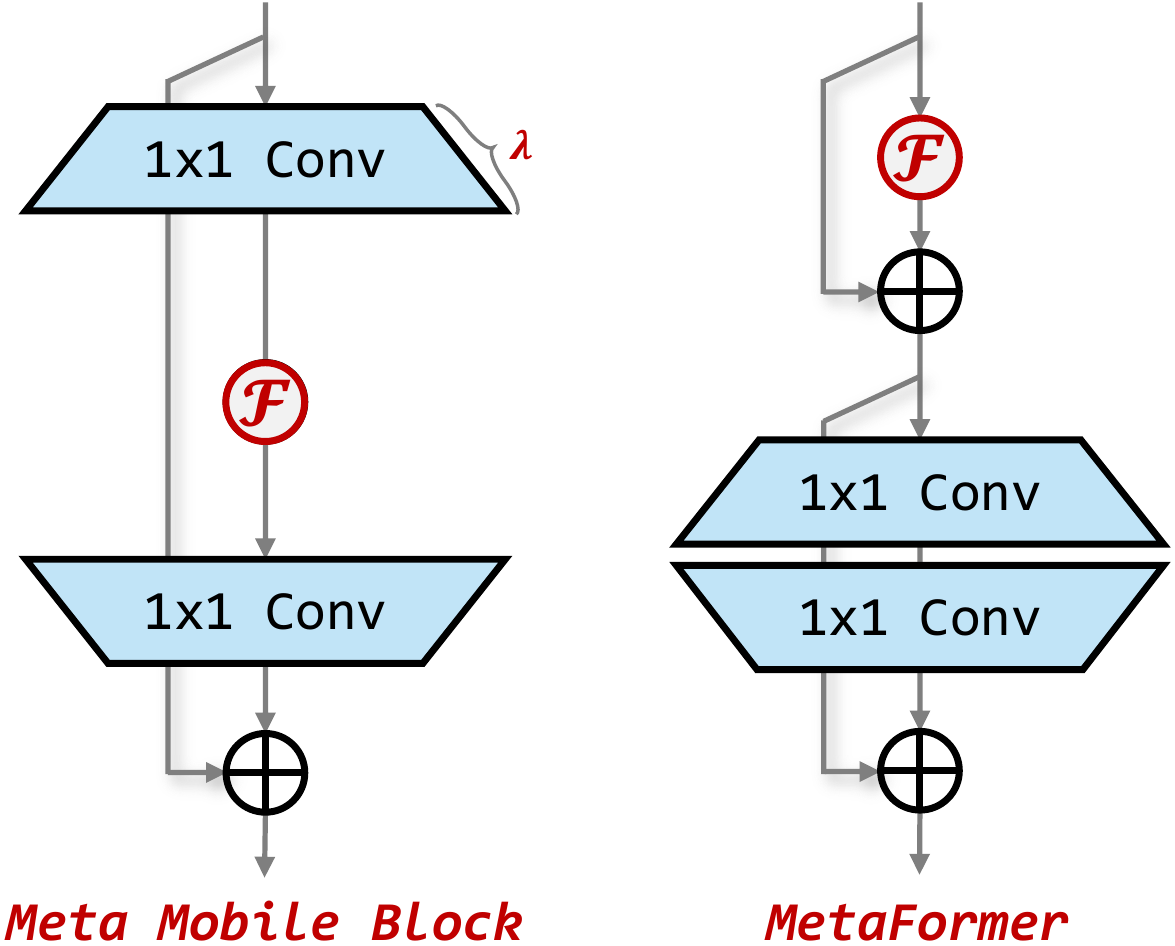}
    \caption{Paradigm illustration with MetaFormer.}
    \vspace{-1.0em}
    \label{fig:com_meta}
\end{wrapfigure}
\noindent\textbf{Relation to MetaFormer.}
We discuss the differences between our \emph{Meta Mobile Block} and \emph{MetaFormer}~\cite{metaformer} in Fig.~\ref{fig:com_meta}. 
\textbf{\emph{1)}} From the structure, two-residual MetaFormer contains two sub-modules with two skip connections, while our Meta Mobile Block contains only one sub-module that covers one-residual IRB in the field of lightweight CNN. Also, shallower depths require less memory access and save costs~\cite{shufflenetv2} that is more general and hardware friendly. 
\textbf{\emph{2)}} From the motivation, MetaFormer is the induction of high-performance Transformer/MLP-like models, while our Meta Mobile Block is the induction of efficient IRB in MobileNetv2~\cite{mnetv2} and effective MHSA/FFN in Transformer~\cite{transformer,vit} for designing efficient infrastructure.
\textbf{\emph{3)}} To a certain extent, the inductive one-residual Meta Mobile Block can be regarded as a conceptual extension of two-residual MetaFormer in the lightweight field. We hope our work inspires more future research dedicated to lightweight model design domain based on attention.

\begin{wrapfigure}{r}{2.8cm}
    \centering
    % \vspace{-1.2em}
    \includegraphics[width=1.0\linewidth]{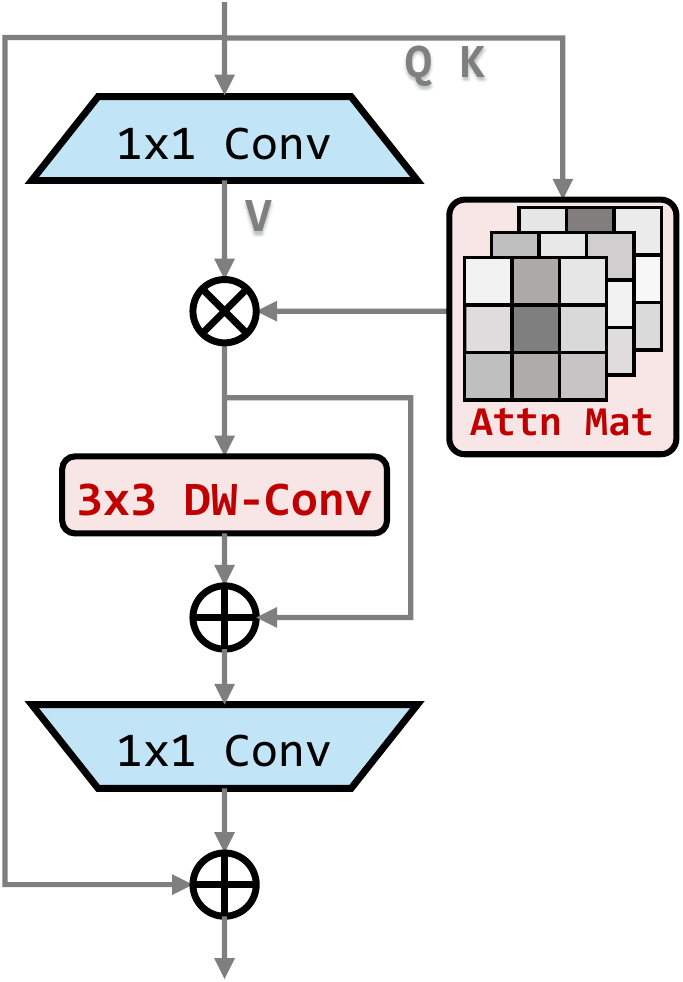}
    \caption{Paradigm of iRMB.}
    \label{fig:irmb}
    \vspace{-1.0em}
\end{wrapfigure}

\begin{table}[b!]
    \centering
    \caption{Complexity and Maximum Path Length analysis of modules. Input/output feature maps are in $\mathbb{R}^{C \times W \times W}$, $L = W^2$, $l = w^2$, $W$ and $w$ are feature map size and window size, while $k$ and $G$ are kernel size and group number.}
    \label{table:mpl}
    \renewcommand{\arraystretch}{1.0}
    \resizebox{1.0\linewidth}{!}{
        \begin{tabular}{p{1.5cm}<{\raggedleft} p{2.1cm}<{\raggedright} p{2.9cm}<{\raggedright} p{2.2cm}<{\raggedright}}
        \toprule[0.17em]
        Module & \#Params & FLOPs & MPL \\
        \hline
        \rowcolor{whit_tab} MHSA   & $4(C+1)C$       & $8C^2L+4CL^2+3L^2$  & $O(1)$    \\
        \rowcolor{whit_tab} W-MHSA   & $4(C+1)C$       & $8C^2L+4CLl+3Ll$  & $O(Inf)$    \\
        \hline
        \rowcolor{whit_tab} Conv  & $(Ck^{2}/G+1)C$ & $(2Ck^{2}/G)LC$     & $O(2W/(k-1))$  \\
        \rowcolor{whit_tab} DW-Conv  & $(k^{2}+1)C$ & $(2k^{2})LC$     & $O(2W/(k-1))$  \\
        \toprule[0.12em]
        \end{tabular}
    }
    % \vspace{-1.0em}
\end{table}

\subsection{Micro Design: Inverted Residual Mobile Block} \label{section:irmb}
Based on the inductive Meta Mobile Block, we instantiate an effective yet efficient modern \emph{Inverted Residual Mobile Block} (iRMB) from a microscopic view in Fig.~\ref{fig:irmb}.\\
\noindent\textbf{Design Principle.} Following criteria in Sec.~\ref{section:cri}, $\deepred{\bm{\mathcal{F}}}$ in iRMB is modeled as cascaded \emph{MHSA} and \emph{Convolution} operations, formulated as $\deepred{\bm{\mathcal{F}}}(\cdot) = \text{Conv}(\text{MHSA}(\cdot))$. This design absorbs CNN-like efficiency to model local features and Transformer-like dynamic modelling capability to learn long-distance interactions. However, naive implementation can lead to unaffordable expenses for two main reasons: \\
\noindent\textit{1)} \deepred{\emph{\bm{$\lambda$}}} is generally greater than one that the intermediate dimension would be multiple to input dimension, causing quadratic \deepred{\emph{\bm{$\lambda$}}} increasing of parameters and computations. Therefore, components of $\deepred{\bm{\mathcal{F}}}$ should be independent or linearly dependent on the number of channels. \\
\noindent\textit{2)} FLOPs of MHSA is proportional to the quadratic of total image pixels, so the cost of a naive Transformer is unaffordable. The specific influences can be seen in Tab.~\ref{table:mpl}. \\
\noindent\textbf{Deduction.} We employ efficient Window-MHSA (W-MHSA) and Depth-Wise Convolution (DW-Conv) with a skip connection to trade-off model cost and accuracy. \\
\noindent\textbf{Improved EW-MHSA.} Parameters and FLOPs for obtaining $Q$,$K$ in W-MHSA is quadratic of the channel, so we employ unexpanded $\boldsymbol{X}$ to calculate the attention matrix more efficiently, \ie, $Q$=$K$=$\boldsymbol{X}$ $(\in\mathbb{R}^{C \times H \times W})$, while the expanded value $\boldsymbol{X}_{e}$ as $V$ $(\in\mathbb{R}^{\deepred{\emph{\bm{$\lambda$}}}C \times H \times W})$. This improvement is termed as \emph{Expanded Window MHSA} (EW-MHSA) that is more applicative, formulated as: 

\begin{equation}
    \begin{aligned}
        \deepred{\bm{\mathcal{F}}}(\cdot) = (\text{DW-Conv, Skip})(\text{EW-MHSA}(\cdot)) \text{.}
    \end{aligned}
\end{equation}
Also, this cascading manner can increase the expansion speed of the receptive field and reduce the maximum path length of the model to $O(2W/(k-1+2w))$, which has been experimentally verified with consistency in Sec.~\ref{section:exp_ablation}.

\noindent\textbf{Flexibility.} Empirically, current transformer-based methods~\cite{edgenext,uniformer,moat} reach a consensus that inductive CNN in shallow layers while global Transformer in deep layers composition could benefit the performance. Unlike recent EdgeNeXt that employs different blocks for different depths, our iRMB satisfies the above design principle using only two switches to control whether two modules are used (Code level is also concise in \#Supp).

\noindent\textbf{Efficient Equivalent Implementation.} MHSA is usually used in channel-consistent projection (\deepred{\emph{\bm{$\lambda$}}}=1), meaning that the FLOPs of multiplying attention matrix times expended $\boldsymbol{X}_{e}$ (\deepred{\emph{\bm{$\lambda$}}}>1) will increase by \deepred{\emph{\bm{$\lambda$}}} - 1. Fortunately, the information flow from $\boldsymbol{X}$ to expended $V$ ($\boldsymbol{X}_{e}$) involves only linear operations, \ie, $\text{MLP}_{e}(\cdot)$, so we can derive an equivalent proposition:"\emph{When the groups of $\text{MLP}_{e}$ equals to the head number of $\text{W-MHSA}$, the multiplication result of exchanging order remains unchanged}." To reduce FLOPs, matrix multiplication before $\text{MLP}_{e}$ is used by default.

\noindent\textbf{Choice of Efficient Operators.} We also replace the component of $\deepred{\bm{\mathcal{F}}}$ with group convolution, asymmetric~\cite{inceptionv3} convolution, and performer~\cite{performer}, but they make no further improvements with much higher parameters and FLOPs at the same magnitude for our approach.

\noindent\textbf{Boosting Naive Transformer.} To assess iRMB performance, we set \deepred{\emph{\bm{$\lambda$}}} to 4 and replace standard Transformer structure in columnar DeiT~\cite{deit} and pyramid-like PVT~\cite{pvtv1}. As shown in Tab.~\ref{table:toy_irmb}, we surprisingly found that iRMB can improve performance with fewer parameters and computations in the same training setting, especially for the columnar ViT. This proves that the one-residual iRMB has obvious advantages over the two-residual Transformer in the lightweight model.

\begin{table}[tp]
    \centering
    \caption{A toy experiment for assessing iRMB.}
    \label{table:toy_irmb}
    \renewcommand{\arraystretch}{1.0}
    \resizebox{1.0\linewidth}{!}{
        \begin{tabular}{p{3.0cm}<{\raggedright} p{1.8cm}<{\raggedright} p{1.8cm}<{\raggedright} p{1.6cm}<{\raggedright}}
        \toprule[0.17em]
        Model & \#Params $\downarrow$ & FLOPs $\downarrow$ & Top-1 $\uparrow$  \\
        \hline
        \rowcolor{whit_tab} DeiT-Tiny~\cite{deit}   & \pzo5.7M                                           & 1258                                          & 72.2  \\
        \rowcolor{lblu_tab} DeiT-Tiny w/iRMB        & \pzo4.9M\scriptsize{\red{${~\text{-}14\%\downarrow}$}}     & 1102\scriptsize{\red{${~\text{-}156M\downarrow}$}}    & 74.3\scriptsize{\red{${~\text{+}2.1\%\uparrow}$}}  \\
        \hline
        \rowcolor{whit_tab} PVT-Tiny~\cite{pvtv1}   & 13.2M                                          & 1943                                          & 75.1  \\
        \rowcolor{lblu_tab} PVT-Tiny w/iRMB         & 11.7M\scriptsize{\red{${~\text{-}11\%\downarrow}$}}    & 1845\scriptsize{\red{${\pzo~\text{-}98M\downarrow}$}}     & 75.4\scriptsize{\red{${~\text{+}0.3\%\uparrow}$}}  \\
        \toprule[0.12em]
        \end{tabular}
    }
    \vspace{-1.0em}
\end{table}

\noindent\textbf{Parallel Design of $\deepred{\bm{\mathcal{F}}}$.} We also implement the parallel structure of DW-Conv and EW-MHSA with half the number of channels in each component, and some configuration details are adaptively modified to ensure the same magnitude. Comparably, this parallel model gets 78.1 (-0.3$\downarrow$) Top-1 in ImageNet-1k dataset with 5.1M parameters and 964M FLOPs (+63M$\uparrow$ than EMO-5M), but its throughput will slow down by about -7\%$\downarrow$. This phenomenon is also discussed in the work~\cite{shufflenetv2} that: "Network fragmentation reduces the degree of parallelism".

\subsection{Macro Design of EMO for Dense Prediction} \label{section:emo} 

Based on the above criteria, we design a ResNet-like 4-phase Efficient MOdel (EMO) based on a series of iRMBs for dense applications, as shown in Fig.~\ref{fig:emo}-\textbf{Right}. \\
\noindent\textbf{\emph{1)}} For the overall framework, EMO consists of only iRMBs without diversified modules$^{\textbf{\ding{193}}}$, which is a departure from recent efficient methods~\cite{mvitv1,edgenext} in terms of designing idea. \\
\noindent\textbf{\emph{2)}} For the specific module, iRMB consists of only standard convolution and multi-head self-attention without other complex operators$^{\textbf{\ding{192}}}$. Also, benefitted by DW-Conv, iRMB can adapt to down-sampling operation through the stride and does not require any position embeddings for introducing inductive bias to MHSA$^{\textbf{\ding{193}}}$. \\
\noindent\textbf{\emph{3)}} For variant settings, we employ gradually increasing expansion rates and channel numbers, and detailed configurations are shown in Tab.~\ref{table:model_variants}. Results for basic classification and multiple downstream tasks in Sec.~\ref{section:exp} demonstrate the superiority of our EMO over SoTA lightweight methods on magnitudes of 1M, 2M, and 5M$^{\textbf{\ding{194}\ding{195}}}$. \\
\noindent\textbf{Details.} Since MHSA is better suited for modelling semantic features for deeper layers, we only turn it on at stage-3/4 following previous works~\cite{edgenext,uniformer,moat}. Note that this never violates the uniformity criterion, as the shutdown of MHSA was a special case of iRMB structure. To further increase the stability of EMO, BN~\cite{bn}+SiLU~\cite{gelu} are bound to DW-Conv while LN~\cite{ln}+GeLU~\cite{gelu} are bound to EW-MHSA. Also, iRMB is competent for down-sampling operations.

\noindent\textbf{Relation to MetaFormer.}
\textbf{\emph{1)}} From the structure, MetaFormer extended dense prediction model employs an extra patch embedding layer for down-sampling, while our EMO only consists of iRMB.
\textbf{\emph{2)}} From the result, our instantiated EMO-5M (w/ 5.1M \#Params and 903M FLOPs) exceeds instantiated PoolFormer-S12 (w/ 11.9M \#Params and 1,823M FLOPs) by +1.2$\uparrow$, illustrating that a stronger efficient operator makes a advantage.
\textbf{\emph{3)}} We further replace Token Mixer in MetaFormer with $\deepred{\bm{\mathcal{F}}}$ in iRMB and build a 5.3M model \vs our EMO-5M. It only achieves 77.5 Top-1 on ImageNet-1k, \ie, -0.9$\downarrow$ than our model, meaning that our proposed Meta Mobile Block has a better advantage for constructing lightweight models than two-residual MetaFormer.

\begin{table}[tp]
    \centering
    \caption{Core configurations of EMO variants.}
    \label{table:model_variants}
    \renewcommand{\arraystretch}{1.0}
    \resizebox{1.0\linewidth}{!}{
        \begin{tabular}{p{1.7cm}<{\raggedright} p{2.8cm}<{\centering} p{3.0cm}<{\centering} p{3.0cm}<{\centering}}
            \toprule[0.17em]
            Items           & EMO-1M                    & EMO-2M                    & EMO-5M \\
            \hline
            Depth           & [ 2, 2, 8, 3 ]            & [ 3, 3, 9, 3 ]            & [ 3, 3, 9, 3 ] \\
            Emb. Dim.       & [ 32, 48,  80, 168 ]      & [ 32, 48, 120, 200 ]      & [ 48, 72, 160, 288 ] \\
            Exp. Ratio      & [ 2.0, 2.5, 3.0, 3.5 ]    & [ 2.0, 2.5, 3.0, 3.5 ]    & [ 2.0, 3.0, 4.0, 4.0 ] \\
            \toprule[0.12em]
        \end{tabular}
    }
    \vspace{-1.0em}
\end{table}

\begin{wraptable}{r}{4.67cm}
    \centering
    \vspace{-1.36em}
    \caption{Ablation study on components in iRMB.}
    \label{table:components_F}
    \renewcommand{\arraystretch}{1.0}
    \setlength\tabcolsep{3.0pt}
    \resizebox{1.0\linewidth}{!}{
        \begin{tabular}{p{1.6cm}<{\centering} p{1.6cm}<{\centering} p{1.8cm}<{\raggedright}}
        \toprule[0.2em]
        \text{EW-MHSA} & \text{DW-Conv} & \pzo\pzo\pzo Top-1 \\
        \hline
        \rowcolor{whit_tab} \xmarkg &   \xmarkg & \pzo73.5  \\
        \rowcolor{whit_tab} \cmark &   \xmarkg  & \pzo76.6\scriptsize{\red{${~\text{+}3.1\uparrow}$}}  \\
        \rowcolor{whit_tab} \xmarkg  &   \cmark & \pzo77.6\scriptsize{\red{${~\text{+}4.1\uparrow}$}}  \\
        \rowcolor{whit_tab} \cmark  &   \cmark  & \pzo78.4\scriptsize{\red{${~\text{+}4.9\uparrow}$}}  \\
        \toprule[0.2em]
        \end{tabular}
    }
    \vspace{-1.0em}
\end{wraptable}

\noindent\textbf{Importance of Instantiated Efficient Operator.} Our defined \emph{efficient operator} \deepred{\bm{$\mathcal{F}$}} contains two core modules, \ie, EW-MHSA and DW-Conv. In Tab.~\ref{table:components_F}, we conduct an ablation experiment to study the effect of both modules. The first row means that neither EW-MHSA nor DW-Conv is used, \ie, the model is almost composed of MLP layers with several DW-Conv for down-sampling, and \deepred{\bm{$\mathcal{F}$}} degenerates to Identity operation. Surprisingly, this model still produces a respectable result, \ie, 73.5 Top-1. Comparatively, results of the second and third rows demonstrate that each component contributes to the performance, \eg, +3.1$\uparrow$ and +4.1$\uparrow$ when adding DW-Conv and EW-MHSA, respectively. Our model achieves the best result, \ie, 78.4 Top-1, when both components are used. Besides, this experiment illustrates that the specific instantiation of iRMB is very important to model performance.\\
\noindent\textbf{Order of Operators.} Based on EMO-5M, we switch the order of DW-Conv/EW-MHSA and find a slight drop in performance (-0.6$\downarrow$), so EW-MHSA performs first by default.

\section{Experiments} \label{section:exp}

\subsection{Image Classification}
\noindent\textbf{Setting.} Due to various training recipes of SoTA methods~\cite{mnetv3,vit,deit,mvitv1,mvitv2,mocovit,edgenext} that could lead to potentially unfair comparisons (summarized in Tab.~\ref{table:train_recipe}), we employ a weaker training recipe to increase model persuasion and open the source code for subsequent fair comparisons in \#Supp. All experiments are conducted on ImageNet-1K dataset~\cite{imagenet} without extra datasets and pre-trained models. Each model is trained for standard 300 epochs from scratch at 224$\times$224, and AdamW~\cite{adamw} optimizer is employed with betas (0.9, 0.999), weight decay 5e$^{-2}$, learning rate 6e$^{-3}$, and batch size 2,048. We use Cosine scheduler~\cite{cosine} with 20 warmup epochs, Label Smoothing 0.1~\cite{label_smoothing}, stochastic depth~\cite{drop_path}, and RandAugment~\cite{randaugment} during training, while LayerScale~\cite{ls}, Dropout~\cite{dropout}, MixUp~\cite{mixup}, CutMix~\cite{cutmix}, Random Erasing~\cite{random_erasing}, Position Embeddings~\cite{vit}, Token Labeling~\cite{token_labeling}, and Multi-Scale training~\cite{mvitv1} are \emph{disabled}. EMO is implemented by PyTorch~\cite{pytorch}, based on TIMM~\cite{timm}, and trained with 8$\times$V100 GPUs.

\noindent\textbf{Results.} EMO is evaluated with SoTAs on three small scales, and quantitative results are shown in Tab.~\ref{table:cls_imagenet}. Surprisingly, our method obtains the current best results without using complex modules and MobileViTv2-like strong training recipe. For example, the smallest EMO-1M obtains SoTA 71.5 Top-1 that surpasses CNN-based MobileNetv3-L-0.50~\cite{mnetv3} by +2.7$\uparrow$ with nearly half parameters and Transformer-based MobileViTv2-0.5~\cite{mvitv2} by +1.3$\uparrow$ with only 56\% FLOPs. Larger EMO-2M achieves SoTA 75.1 Top-1 with only 439M FLOPs, nearly half of MobileVit-XS~\cite{mvitv1}. Comparatively, the latest EdgeViT-XXX~\cite{edgevit} obtains a worse 74.4 Top-1 while requiring +78\%$\uparrow$ parameters and +27\%$\uparrow$ FLOPs. Consistently, EMO-5M obtains a superior trade-off between \#Params (5.1M) / \#FLOPs (903M) and accuracy (78.4), which is more efficient than contemporary counterparts. Surprisingly, after increasing the channel of the fourth stage of EMO-5M from 288 to 320, the new EMO-6M reaches 79.0 Top-1 with only 961M FLOPs.

\begin{table}[tp]
    \centering
    \caption{Comparison of \textbf{training recipes among contemporary methods} and we employ the same setting in all experiments. Please zoom in for clearer comparisons. Abbreviated MNet and MViT: MobileNet and MobileViT.}
    \label{table:train_recipe}
    \renewcommand{\arraystretch}{1.0}
    \setlength\tabcolsep{2.0pt}
    \resizebox{1.\linewidth}{!}{
        \begin{tabular}{p{2.8cm}<{\centering} p{1.7cm}<{\centering} p{1.7cm}<{\centering} p{1.7cm}<{\centering} p{1.8cm}<{\centering} p{1.8cm}<{\centering} p{1.8cm}<{\centering} p{1.5cm}<{\centering}}
        \toprule
        \multirow{2}{*}{Super-Params.} & \multirow{2}{*}{\makecell[c]{MNetv3~\cite{mnetv3}\\ \grayer{ICCV'19}}} & \multirow{2}{*}{\makecell[c]{ViT~\cite{vit}\\ \grayer{ICLR'21}}} & \multirow{2}{*}{\makecell[c]{DeiT~\cite{deit}\\ \grayer{ICML'21}}} & \multirow{2}{*}{\makecell[c]{MViTv1~\cite{mvitv1}\\ \grayer{ICLR'22}}} & \multirow{2}{*}{\makecell[c]{MViTv2~\cite{mvitv2}\\ \grayer{arXiv'22}}} & \multirow{2}{*}{\makecell[c]{EdgeNeXt~\cite{edgenext}\\ \grayer{arXiv'22}}} & \multirow{2}{*}{\makecell[c]{EMO\\ \deepred{Ours}}} \\
        & & & & & & & \\
        \hline
        Epochs              & 300           & 300       & 300       & 300       & 300       & 300       & 300       \\
        Batch size          & 512           & 4096      & 1024      & 1024      & 1024      & 4096      & 2048      \\
        Optimizer           & RMSprop       & AdamW     & AdamW     & AdamW     & AdamW     & AdamW     & AdamW     \\
        Learning rate       & 6.4e$^{-2}$   & 3e$^{-3}$ & 1e$^{-3}$ & 2e$^{-3}$ & 2e$^{-3}$ & 6e$^{-3}$ & 6e$^{-3}$ \\
        Learning rate decay & 1e$^{-5}$     & 3e$^{-1}$ & 5e$^{-2}$ & 1e$^{-2}$ & 5e$^{-2}$ & 5e$^{-2}$ & 5e$^{-2}$ \\
        Warmup epochs       & 3             & 3.4       & 5         & 2.4       & 16        & 20        & 20        \\
        \hline
        Label smoothing     & 0.1           & \xmark    & 0.1       & 0.1       & 0.1       & 0.1       & 0.1       \\
        Drop out rate       & \xmark        & 0.1       & \xmark    & 0.1       & \xmark    & \xmark    & \xmark    \\
        Drop path rate      & 0.2           & \xmark    & 0.1       & \xmark    & \xmark    & 0.1       & 0.1       \\
        \hline
        RandAugment         & 9/0.5         & \xmark    & 9/0.5     & \xmark    & 9/0.5     & 9/0.5     & 9/0.5     \\
        Mixup alpha         & \xmark        & \xmark    & 0.8       & \xmark    & 0.8       & \xmark    & \xmark    \\
        Cutmix alpha        & \xmark        & \xmark    & 1.0       & \xmark    & 1.0       & \xmark    & \xmark    \\
        Erasing probability & 0.2           & \xmark    & 0.25      & \xmark    & 0.25      & \xmark    & \xmark    \\
        Position embedding  & \xmark        & \cmark    & \cmark    & \xmark    & \xmark    & \cmark    & \xmark    \\
        Multi-scale sampler & \xmark        & \xmark    & \xmark    & \cmark    & \xmark    & \cmark    & \xmark    \\
        \bottomrule
        \end{tabular}
    }
    \vspace{-1.0em}
\end{table}

\begin{table}[htp]
    \centering
    \caption{Classification performance on ImageNet-1K dataset. \protect\sethlcolor{whit_tab}\hl{White}, \protect\sethlcolor{oran_tab}\hl{yellow}, and \protect\sethlcolor{blue_tab}\hl{blue} backgrounds indicate CNN-based, Transformer-based, and our EMO, respectively. This kind of display continues for all subsequent experiments. Unit: (M). Abbreviated MNet and MViT: MobileNet and MobileViT.} 
    \label{table:cls_imagenet}
    \renewcommand{\arraystretch}{1.0}
    \setlength\tabcolsep{5.0pt}
    \resizebox{1.\linewidth}{!}{
        \begin{tabular}{p{3.0cm}<{\raggedright} p{1.5cm}<{\centering} p{1.36cm}<{\centering} p{0.76cm}<{\centering} p{0.9cm}<{\centering} p{1.5cm}<{\raggedleft}}
        \toprule[0.17em]
        Model & \#Params $\downarrow$ & FLOPs $\downarrow$ & Reso. & Top-1 & \#Pub\pzo\pzo \\
        \hline
        % \rowcolor{whit_tab} MNetv1-0.50~\cite{mnetv1}               & 1.3   & 149   & $224^{2}$ & 63.7  & arXiv'17  \\
        \rowcolor{whit_tab} MNetv3-L-0.50~\cite{mnetv3}             & 2.6   & 69    & $224^{2}$ & 68.8  & ICCV'19   \\
        \rowcolor{oran_tab} MViTv1-XXS~\cite{mvitv1}                & 1.3   & 364   & $256^{2}$ & 69.0  & ICLR'22   \\
        \rowcolor{oran_tab} MViTv2-0.5~\cite{mvitv2}                & 1.4   & 466   & $256^{2}$ & 70.2  & arXiv'22  \\
        \rowcolor{oran_tab} EdgeNeXt-XXS~\cite{edgenext}            & 1.3   & 261   & $256^{2}$ & 71.2  & ECCVW'22  \\
        \rowcolor{blue_tab} \textbf{EMO-1M}                         & 1.3   & 261   & $224^{2}$ & 71.5  &  ICCV'23       \\
        \hline
        \hline
        \rowcolor{whit_tab} MNetv2-1.40~\cite{mnetv2}               & 6.9   & 585   & $224^{2}$ & 74.7  & CVPR'18   \\
        \rowcolor{whit_tab} MNetv3-L-0.75~\cite{mnetv3}             & 4.0   & 155   & $224^{2}$ & 73.3  & ICCV'19   \\
        \rowcolor{oran_tab} MoCoViT-1.0~\cite{mocovit}              & 5.3   & 147   & $224^{2}$ & 74.5 & arXiv'22   \\
        \rowcolor{oran_tab} PVTv2-B0~\cite{pvtv2}                   & 3.7   & 572   & $224^{2}$ & 70.5 & CVM'22   \\
        \rowcolor{oran_tab} MViTv1-XS~\cite{mvitv1}                 & 2.3   & 986   & $256^{2}$ & 74.8 & ICLR'22\\
        \rowcolor{oran_tab} MFormer-96M~\cite{mobileformer}         & 4.6   & 96    & $224^{2}$ & 72.8 & CVPR'22   \\
        \rowcolor{oran_tab} EdgeNeXt-XS~\cite{edgenext}             & 2.3   & 538   & $256^{2}$ & 75.0 & ECCVW'22\\
        \rowcolor{oran_tab} EdgeViT-XXS~\cite{edgevit}              & 4.1   & 557   & $256^{2}$ & 74.4 & ECCV'22\\
        \rowcolor{oran_tab} tiny-MOAT-0~\cite{moat}              & 3.4   & 800   & $224^{2}$ & 75.5 & ICLR'23\\
        \rowcolor{blue_tab} \textbf{EMO-2M}                         & 2.3   & 439   & $224^{2}$ & 75.1 & ICCV'23   \\
        \hline
        \hline
        \rowcolor{whit_tab} MNetv3-L-1.25~\cite{mnetv3}             & 7.5   & 356   & $224^{2}$ & 76.6 & ICCV'19   \\
        \rowcolor{whit_tab} EfficientNet-B0~\cite{efficientnet}     & 5.3   & 399   & $224^{2}$ & 77.1 & ICML'19 \\
        \rowcolor{oran_tab} DeiT-Ti~\cite{deit}                     & 5.7   & 1258  & $224^{2}$ & 72.2 & ICML'21 \\
        \rowcolor{oran_tab} XCiT-T12~\cite{xcit}                    & 6.7   & 1254  & $224^{2}$ & 77.1 & NIPS'21\\
        \rowcolor{oran_tab} LightViT-T~\cite{lightvit}              & 9.4   & 700   & $224^{2}$ & 78.7 & arXiv'22\\
        \rowcolor{oran_tab} MViTv1-S~\cite{mvitv1}                  & 5.6   & 2009  & $256^{2}$ & 78.4 & ICLR'22\\
        \rowcolor{oran_tab} MViTv2-1.0~\cite{mvitv2}                & 4.9   & 1851  & $256^{2}$ & 78.1 & arXiv'22\\
        \rowcolor{oran_tab} EdgeNeXt-S~\cite{edgenext}              & 5.6   & 965   & $224^{2}$ & 78.8 & ECCVW'22\\
        \rowcolor{oran_tab} PoolFormer-S12~\cite{metaformer}        & 11.9  & 1823  & $224^{2}$ & 77.2 & CVPR'22 \\
        \rowcolor{oran_tab} MFormer-294M~\cite{mobileformer}        & 11.4  & 294   & $224^{2}$ & 77.9 & CVPR'22 \\
        \rowcolor{oran_tab} MPViT-T~\cite{mpvit}                    & 5.8   & 1654  & $224^{2}$ & 78.2 & CVPR'22\\
        \rowcolor{oran_tab} EdgeViT-XS~\cite{edgevit}               & 6.7   & 1136  & $256^{2}$ & 77.5 & ECCV'22\\
        \rowcolor{oran_tab} tiny-MOAT-1~\cite{moat}              & 5.1   & 1200  & $224^{2}$ & 78.3 & ICLR'23\\
        \rowcolor{blue_tab} \textbf{EMO-5M}                         & 5.1   & 903   & $224^{2}$ & 78.4 & ICCV'23   \\
        \rowcolor{blue_tab} \textbf{EMO-6M}                         & 6.1   & 961   & $224^{2}$ & 79.0 &  ICCV'23   \\
        \toprule[0.12em]
        \end{tabular}
    }
    \vspace{-1.5em}
\end{table}

\noindent\textbf{Training Recipes Matters.} We evaluate EMO with different training recipes: 
\begin{center}\vspace{-.2em}
\tablestyle{6pt}{1.05}
\setlength\tabcolsep{1.0pt}
\begin{tabular}{p{1.5cm}<{\centering}p{1.5cm}<{\centering}p{1.5cm}<{\centering}p{1.5cm}<{\centering}}
MNetv3 & DeiT & EdgeNeXt & EMO \\
\shline
NaN & 78.1 & 78.3 & 78.4
\end{tabular}\vspace{-.2em}
\end{center}
We find that the simple training recipe (Ours) is enough to get good results for our lightweight EMO, while existing stronger recipes (especially in EdgeNeXt~\cite{edgenext}) will not improve the model further. NaN indicates the model did not train well for the possibly unadapted hyper-parameters.

\subsection{Downstream Tasks}
\noindent\textbf{Object detection.} ImageNet-1K pre-trained EMO is integrated with light SSDLite~\cite{mnetv3} and heavy RetinaNet~\cite{retinanet} to evaluate its performance on MS-COCO 2017~\cite{coco} dataset at 320$\times$320 resolution. Considering fairness and friendliness for the community, we employ standard MMDetection library~\cite{mmdetection} for experiments and replace the optimizer with AdamW~\cite{adamw} without tuning other parameters. 

For SSDLite, comparison results with SoTA methods are shown in Tab.~\ref{table:det_coco_ssdlite}, and our EMO surpasses corresponding counterparts by apparent advantages. For example, SSDLite equipped with EMO-1M achieves 22.0 mAP with only 0.6G FLOPs and 2.3M parameters, which boosts +2.1$\uparrow$ compared with SoTA MobileViT~\cite{mvitv1} with only 66\% FLOPs. Consistently, our EMO-5M obtains the highest 27.9 mAP so far with much fewer FLOPs, \eg, 53\% (1.8G) of MobileViT-S~\cite{mvitv1} (3.4G) and 0.3G less than EdgeNeXt-S (2.1G). For RetinaNet, data in Tab.~\ref{table:det_coco_retinanet} come from official EdgeViT~\cite{edgevit}, and our EMO consistently obtains better results over counterparts, \eg, +2.6$\uparrow$ AP than CNN-based ResNet-50 and +1.7$\uparrow$ AP than Transformer-based PVTv2-B0. In addition, we report EMO-5M-based RetinaNet with 178.11 GFLOPs for the follow-up comparison.

\begin{figure}[tp]
    \centering
    \includegraphics[width=1.0\linewidth]{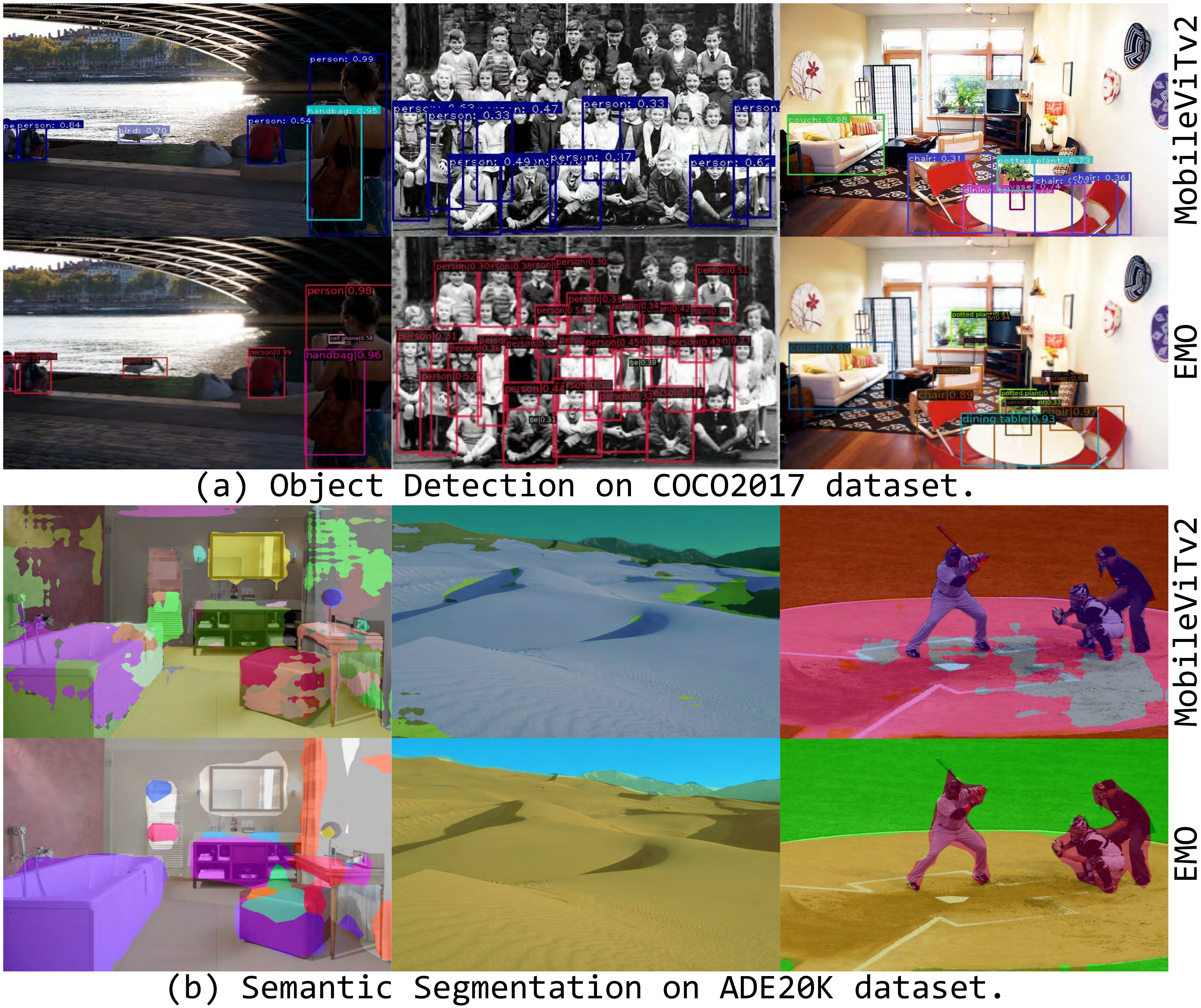}
    \caption{Qualitative comparisons with MobileNetv2 on two main downstream tasks. Zoom in for more details.}
    \label{fig:det_seg}
\end{figure}

Qualitative detection visualizations compared with MobileViTv2 by SSDLite are shown in Fig.~\ref{fig:det_seg}-(a), and results indicate the superiority of our EMO for capturing adequate and accurate information on different scenes. 

\begin{table}[tp]
    \centering
    \caption{Object detection performance by SSDLite on MS-COCO. Abbreviated MNet/MViT: MobileNet/MobileViT.}
    \label{table:det_coco_ssdlite}
    \renewcommand{\arraystretch}{1.0}
    \setlength\tabcolsep{8.0pt}
    \resizebox{1.0\linewidth}{!}{
        \begin{tabular}{p{2.7cm}<{\raggedright} p{1.6cm}<{\centering} p{1.6cm}<{\centering} p{1.6cm}<{\centering}}
        \toprule[0.17em]
        Backbone & \#Params $\downarrow$ & FLOPs $\downarrow$ & mAP \\
        \hline
        \rowcolor{whit_tab} MNetv1~\cite{mnetv1}        & 5.1       & 1.3G      & 22.2 \\
        \rowcolor{whit_tab} MNetv2~\cite{mnetv2}        & 4.3       & 0.8G      & 22.1 \\
        \rowcolor{whit_tab} MNetv3~\cite{mnetv3}        & 5.0       & 0.6G      & 22.0 \\
        \hline
        \hline
        \rowcolor{oran_tab} MViTv1-XXS~\cite{mvitv1}    & 1.7       & 0.9G      & 19.9 \\
        \rowcolor{oran_tab} MViTv2-0.5~\cite{mvitv2}    & 2.0       & 0.9G      & 21.2 \\
        \rowcolor{blue_tab} \textbf{EMO-1M}             & 2.3       & 0.6G      & 22.0 \\
        \hline
        \hline
        \rowcolor{oran_tab} MViTv2-0.75~\cite{mvitv2}   & 3.6       & 1.8G      & 24.6 \\
        \rowcolor{blue_tab} \textbf{EMO-2M}             & 3.3       & 0.9G      & 25.2\\
        \hline
        \hline
        \rowcolor{whit_tab} ResNet50~\cite{resnet}      & 26.6      & 8.8G      & 25.2 \\
        \rowcolor{oran_tab} MViTv1-S~\cite{mvitv1}      & 5.7       & 3.4G      & 27.7 \\
        \rowcolor{oran_tab} MViTv2-1.25~\cite{mvitv2}   & 8.2       & 4.7G      & 27.8 \\
        \rowcolor{oran_tab} EdgeNeXt-S~\cite{edgenext}  & 6.2       & 2.1G      & 27.9 \\
        \rowcolor{blue_tab} \textbf{EMO-5M}             & 6.0       & 1.8G      & 27.9 \\
        \toprule[0.12em]
        \end{tabular}
    }
    \vspace{-1.5em}
\end{table}

\begin{table}[tp]
    \centering
    \caption{Object detection results by RetinaNet on MS-COCO.}
    \label{table:det_coco_retinanet}
    \renewcommand{\arraystretch}{1.1}
    \setlength\tabcolsep{1.0pt}
    \resizebox{1.\linewidth}{!}{
        \begin{tabular}{p{2.8cm}<{\raggedright} p{1.36cm}<{\centering} p{1.0cm}<{\centering} p{1.0cm}<{\centering} p{1.0cm}<{\centering} p{1.0cm}<{\centering} p{1.0cm}<{\centering} p{1.0cm}<{\centering}}
        \toprule[0.17em]
        Backbone & \#Params & AP & AP$_{50}$ & AP$_{75}$ & AP$_{S}$ & AP$_{M}$ & AP$_{L}$ \\
        \hline
        \rowcolor{whit_tab} ResNet-50~\cite{resnet}     & 37.7 & 36.3 & 55.3 & 38.6 & 19.3 & 40.0 & 48.8 \\
        \rowcolor{oran_tab} PVTv1-Tiny~\cite{pvtv1}     & 23.0 & 36.7 & 56.9 & 38.9 & 22.6 & 38.8 & 50.0 \\
        \rowcolor{oran_tab} PVTv2-B0~\cite{pvtv2}       & 13.0 & 37.2 & 57.2 & 39.5 & 23.1 & 40.4 & 49.7 \\
        % \rowcolor{oran_tab} EdgeViT-XXS~\cite{edgevit}  & 13.1 & 38.7 & 59.0 & 41.0 & 22.4 & 42.0 & 51.6 \\
        \rowcolor{blue_tab} EMO-5M                      & 14.4 & 38.9 & 59.8 & 41.0 & 23.8 & 42.2 & 51.7 \\
        \toprule[0.12em]
        \end{tabular}
    }
    \vspace{-1.0em}
\end{table}

\noindent\textbf{Semantic segmentation.} ImageNet-1K pre-trained EMO is integrated with DeepLabv3~\cite{deeplabv3} and PSPNet~\cite{pspnet} to adequately evaluate its performance on challenging ADE20K~\cite{ade20k} dataset at 512$\times$512 resolution. Also, we employ standard MMSegmentation library~\cite{mmseg2020} for experiments and replace the optimizer with AdamW~\cite{adamw} without tuning other parameters. Details can be viewed in the code.

Comparison results with SoTA methods are shown in Tab.~\ref{table:seg_ade_deeplabv3_pspnet}, and our EMO is apparently superior over SoTA Transformer-based MobileViTv2~\cite{mvitv2} at various scales when integrating into segmentation frameworks. For example, EMO-1M/2M/5M armed DeepLabv3 obtains 33.5/35.3/37.8 mIoU, surpassing MobileViTv2 counterparts by +1.6$\uparrow$/+0.6$\uparrow$/+0.6$\uparrow$, while owning fewer parameters and FLOPs benefitted from efficient iRMB. Also, consistent conclusions can be reached when applying EMO as the backbone network of PSPNet. More qualitative results in \#Supp.

Qualitative segmentation results compared with MobileViTv2 by DeepLabv3 are shown in Fig.~\ref{fig:det_seg}-(b), and EMO-based model can obtain more accurate and stable results than the comparison approach, \eg, more consistent bathtub, sand, and baseball field segmentation results.

\begin{table}[tp]
    \centering
    \caption{Semantic segmentation performance on ADE20K dataset. Abbreviated MNet/MViT: MobileNet/MobileViT.}
    \label{table:seg_ade_deeplabv3_pspnet}
    \renewcommand{\arraystretch}{1.0}
    \setlength\tabcolsep{1.0pt}
    \resizebox{1.0\linewidth}{!}{
        \begin{tabular}{p{2.6cm}<{\raggedright} p{1.36cm}<{\centering} p{1.36cm}<{\centering} p{1.36cm}<{\centering} p{1.36cm}<{\centering} p{1.36cm}<{\centering} p{1.36cm}<{\centering}}
        \toprule[0.2em]

        \multirow{2}{*}{Backbone} & \multicolumn{3}{c}{\makecell[c]{DeepLabv3~\cite{deeplabv3}}} & \multicolumn{3}{c}{\makecell[c]{PSPNet~\cite{pspnet}}} \\
        \cmidrule(lr){2-4} \cmidrule(lr){5-7}
        & \#Params & FLOPs & mIoU & \#Params & FLOPs & mIoU \\
        \hline
        \rowcolor{oran_tab} MViTv2-0.5~\cite{mvitv2}    & 6.3   & 26.1  & 31.9  & 3.6   & 15.4  & 31.8 \\
        \rowcolor{blue_tab} \textbf{EMO-1M}             & 5.6   & 2.4   & 33.5  & 4.3   & 2.1   & 33.2 \\
        \hline
        \hline
        \rowcolor{whit_tab} MNetv2~\cite{mnetv2}        & 18.7  & 75.4  & 34.1  & 13.7  & 53.1  & 29.7 \\
        \rowcolor{oran_tab} MViTv2-0.75~\cite{mvitv2}   & 9.6   & 40.0  & 34.7  & 6.2   & 26.6  & 35.2 \\
        \rowcolor{blue_tab} \textbf{EMO-2M}             & 6.9   & 3.5   & 35.3  & 5.5   & 3.1   & 34.5 \\
        \hline
        \hline
        \rowcolor{oran_tab} MViTv2-1.0~\cite{mvitv2}    & 13.4  & 56.4  & 37.0  & 9.4   & 40.3  & 36.5 \\
        \rowcolor{blue_tab} \textbf{EMO-5M}             & 10.3  & 5.8   & 37.8  & 8.5   & 5.3   & 38.2 \\
        \toprule[0.2em]
        \end{tabular}
    }
    \vspace{-1.5em}
\end{table}

\subsection{Extra Ablation and Explanatory Analysis} \label{section:exp_ablation}

\noindent\textbf{Throughput Comparison.} In Tab.~\ref{table:exp_throughput}, we present throughput evaluation results compared with SoTA EdgeNeXt~\cite{edgenext}. The test platforms are AMD EPYC 7K62 CPU and V100 GPU with a resolution of 224$\times$224 and a batch size of 256. Results indicate that our EMO has an faster speed on both platforms, even though both methods have similar FLOPs. For example, EMO-1M achieves speed boosts of +20\%$\uparrow$ for GPU and +116\%$\uparrow$ for CPU than EdgeNeXt-XXS over the same FLOPs. This gap is further widened on mobile devices (following official classification project~\cite{ios_cls} by iPhone14), \ie, \textit{2.8$\times\uparrow$}, \textit{3.9$\times\uparrow$}, and \textit{4.80$\times\uparrow$} faster than SoAT EdgeNeXt~\cite{edgenext}. This derives from our simple and device-friendly iRMB with no other complex structures, \eg, Res2Net module~\cite{res2net}, transposed channel attention~\cite{xcit}, \etc. 

\begin{table}[tp]
    \centering
    \caption{Comparisons of throughput on CPU/GPU and running speed on mobile iPhone14 (ms).}
    \label{table:exp_throughput}
    \renewcommand{\arraystretch}{1.0}
    \setlength\tabcolsep{3.0pt}
    \resizebox{1.0\linewidth}{!}{
        \begin{tabular}{p{2.4cm}<{\raggedright} p{1.2cm}<{\centering} p{1.2cm}<{\centering} p{1.2cm}<{\centering} p{1.8cm}<{\raggedright}}
        \toprule[0.2em]
        \pzo\pzo Method & FLOPs & CPU & GPU & iPhone14\\
        \hline
        \rowcolor{oran_tab} EdgeNeXt-XXS    & 261M & \pzo73.1   & 2860.6    & 12.6 \\
        \rowcolor{blue_tab} \textbf{EMO-1M} & 261M & 158.4  & 3414.6    & \pzo4.5\scriptsize{\red{${~2.8\times\uparrow}$}} \\
        \hline
        \rowcolor{whit_tab} EdgeNeXt-XS     & 538M & \pzo69.1   & 1855.2    & 20.2 \\
        \rowcolor{blue_tab} \textbf{EMO-2M} & 439M & 126.6  & 2509.8    & \pzo5.1\scriptsize{\red{${~3.9\times\uparrow}$}} \\
        \hline
        \rowcolor{oran_tab} EdgeNeXt-S      & 965M & \pzo54.2   & 1622.5    & 27.7 \\
        \rowcolor{blue_tab} \textbf{EMO-5M} & 903M & 106.5  & 1731.7    & \pzo6.8\scriptsize{\red{${~4.0\times\uparrow}$}} \\
        \toprule[0.2em]
        \end{tabular}
    }
    \vspace{-1.5em}
\end{table}

\begin{table}[tp]
    \centering
    \caption{Performance \vs depth configurations.}
    \label{table:exp_depth}
    \renewcommand{\arraystretch}{1.0}
    \setlength\tabcolsep{5.0pt}
    \resizebox{0.9\linewidth}{!}{
        \begin{tabular}{p{2.3cm}<{\centering} p{1.7cm}<{\centering} p{1.7cm}<{\centering} p{1.7cm}<{\centering}}
        \toprule[0.2em]
        $~$Depth & \#Params & FLOPs & Top-1 \\
        \hline
        $~$[ 2, 2, 10, 3 ] & 5.3M & 901M & 78.0 \\
        $~$[ 2, 2, 12, 2 ] & 5.0M & 970M & 77.8 \\
        $~$[ 4, 4, \pzo8, 3 ] & 4.9M & 905M & 78.1 \\
        $~$[ 3, 3, \pzo9, 3 ] & 5.1M & 903M & 78.4 \\
        \toprule[0.2em]
        \end{tabular}
    }
    \vspace{-1.5em}
\end{table}

\noindent\textbf{Depth Configuration.} We assess another three models with different depths on the order of 5M in Tab.~\ref{table:exp_depth}. The selected depth configuration produces relatively better performance.

\noindent\textbf{Comparison with EfficientNet/EfficientFormer.} The manually designed EMO trade-offs \#Params, FLOPs, and performance, and \#Params lies in 1M/2M/5M scales, thus NAS-assisted EfficientNet-B1 (ENet-B1)~\cite{efficientnet} with 7.8M and EfficientFormer-L1 (EFormer-L1)~\cite{efficientformer} with 12.3M are not included in Tab.~\ref{table:cls_imagenet}. Comparatively, our EMO-6M obtains a competitive 79.0 Top-1 with much less \#Params over them, arguing that EMO achieves a better trade-off among \#Params, FLOPs, and performance. Also, our \textit{roughly manually designed model} is promising for further performance improvements with more rational configurations in future work. Benefit from EW-MHSA, EMO offers clear advantages for high-resolution downstream tasks, \eg, compared with more powerful EfficientNet-B2 (ENet-B2) and EfficientFormer-L1 with higher \#Params/FLOPs, EMO-5M achieves better performance as belows:
\vspace{-1.5em}
\begin{center}
    \tablestyle{3pt}{1.05}
    \renewcommand{\arraystretch}{1.1}
    \setlength\tabcolsep{1.0pt}
    \resizebox{1.0\linewidth}{!}{
        \begin{tabular}{p{1.9cm}<{\centering} p{1.7cm}<{\centering} p{1.9cm}<{\centering} p{1.9cm}<{\centering} p{1.7cm}<{\centering} p{1.7cm}<{\centering}}
        % EMO-5M & 1 & 1 & 1 & 1 & 1 \\
        \multirow{2}{*}{Backbone} & \multirow{2}{*}{\makecell[c]{mAP\\SSDLite}} & \multirow{2}{*}{\makecell[c]{AP$^{box}$\\Mask RCNN}} & \multirow{2}{*}{\makecell[c]{AP$^{mask}$\\Mask RCNN}} & \multirow{2}{*}{\makecell[c]{mIoU\\DeepLabv3}} & \multirow{2}{*}{\makecell[c]{mIoU\\Semantic FPN}} \\
        & & & & & \\
        \hline
        ENet-B1 & 27.3 & 38.0 & 35.2 & 36.6 & 38.5  \\
        ENet-B2 & 27.5 & 38.5 & 35.6 & 37.0 & 39.1  \\
        EFormer-L1 & - & 37.9 & 35.4 & - & 38.9 \\
        EMO-5M & \textbf{27.9} & \textbf{39.3} & \textbf{36.4} & \textbf{37.8} & \textbf{40.3} \\
        \end{tabular}
    }
\end{center}
\vspace{-0.5em}

\begin{figure}[tp]
    \centering
    \includegraphics[width=1.0\linewidth]{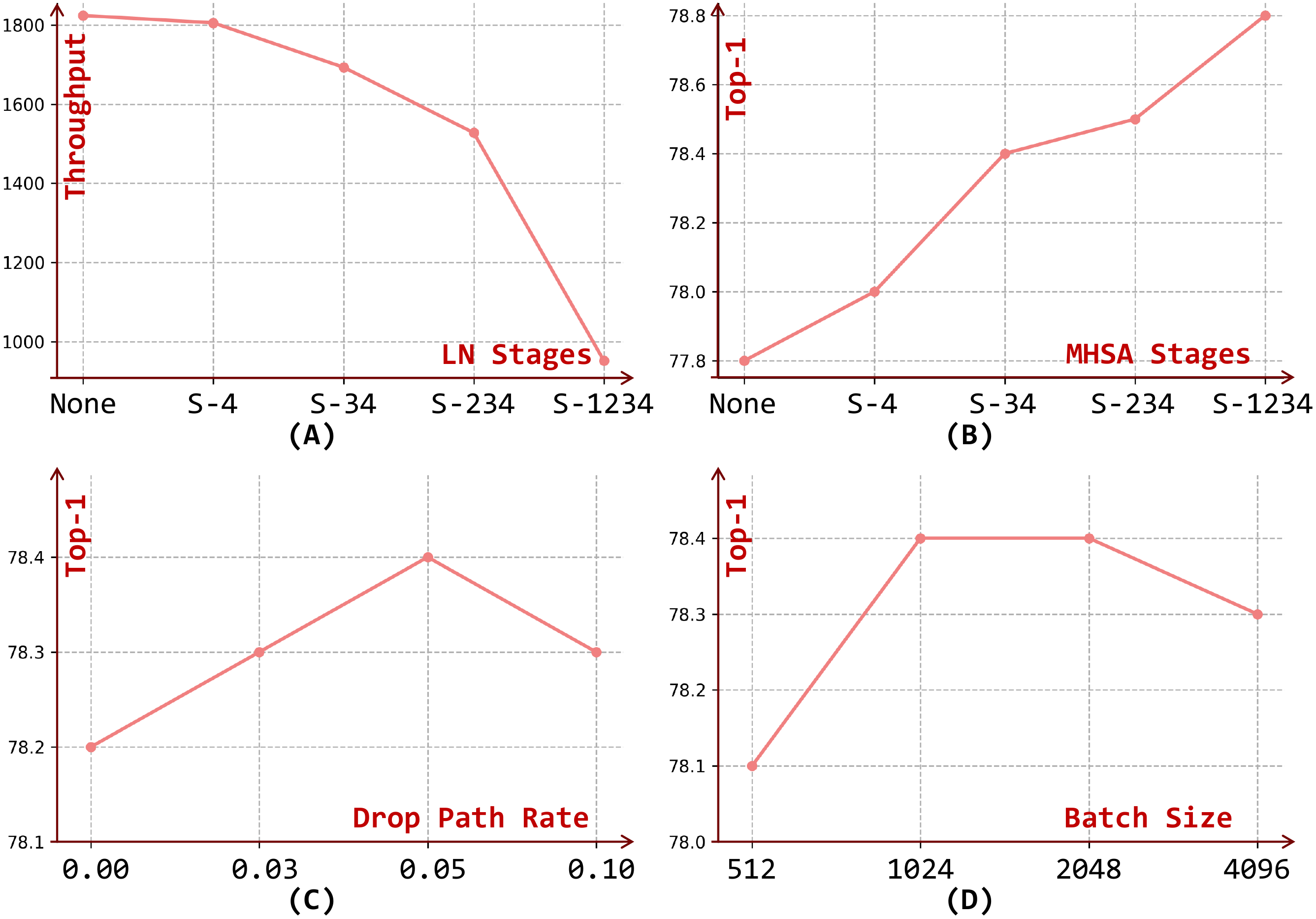}
    \caption{Ablation studies on ImageNet-1K with EMO-5M.}
    \label{fig:ablation}
    \vspace{-1.0em}
\end{figure}

\noindent\textbf{Normalization Type in Different Stages.} BN and LN of the same dimension have the same parameters and similar FLOPs, but LN has a tremendous negative impact on the speed of vision models limited by the underlying optimization of GPU structure. Fig.~\ref{fig:ablation}\red{A} shows the throughput of EMO-5M with the LN layer applying to different stages, and LN is used to stage-3/4 (S-34) by default. As more stages replace BN with LN, \ie, S-1234, throughput decreases significantly (1,693 $\rightarrow$ 952) while the benefit is modest (+0.2$\uparrow$). We found that the model is prone to unstable NaNs when LN is not used; thus, we argue that LN is necessary but used in a few stages is enough for Attention-based EMO.

\noindent\textbf{MHSA in Different Stages.} Fig.~\ref{fig:ablation}\red{B} illustrates the changes in model accuracy when applying MHSA to different stages based on EMO-5M, and we further detail that as follows:
\vspace{-1.0em}
\begin{center}
    \tablestyle{3pt}{1.05}
    \renewcommand{\arraystretch}{1.1}
    \setlength\tabcolsep{2.0pt}
    \resizebox{1.0\linewidth}{!}{
        \begin{tabular}{p{0.6cm}<{\centering} p{0.6cm}<{\centering} p{0.6cm}<{\centering} p{0.6cm}<{\centering} p{1.3cm}<{\centering} p{1.2cm}<{\centering} p{0.9cm}<{\centering} p{0.9cm}<{\centering} p{1.0cm}<{\centering} p{1.2cm}<{\centering}}
        S1 & S2 & S3 & S4 & \#Params & FLOPs & Top-1 & CPU & GPU & iPhone14 \\
        \hline
        % \xmarkg & \xmarkg & \xmarkg & \xmarkg & 4.364 & 778 & 77.8 \\
        \xmarkg & \xmarkg            & \xmarkg   & \cmark & 4.697 & 797 & 78.0 & 205.7    & 2111.4    & 4.9 \\
        \multicolumn{2}{l}{EMO-5M}   & \cmark    & \cmark & 5.109 & 903 & 78.4 & 106.5    & 1731.7    & 6.8\\
        \xmarkg & \cmark             & \cmark    & \cmark & 5.130 & 931 & 78.5 & \pzo98.0 & 1492.2    & 7.5 \\
        \cmark  & \cmark             & \cmark    & \cmark & 5.139 & 992 & 78.8 & \pzo52.3 & \pzo886.8 & 9.5 \\
        \end{tabular}
    }
\end{center}
\vspace{-0.5em}
Results indicate that MHSA always positively affects model accuracy no matter what stage inserted. Our efficient model obtains the best result when applying MHSA to every stage, but this would take an extra 10\%$\uparrow$ more FLOPs, \ie, from 903M to 992M. Therefore, only using MHSA in the last two stages is used by default, which trades off the accuracy and efficiency of the model.

\noindent\textbf{Effect of Drop Path Rate.} Fig.~\ref{fig:ablation}\red{C} explores the effect of drop path rate for training EMO-5M. Results show that the proposed model is robust to this training parameter in the range [0, 0.1] that fluctuates accuracy within 0.2, and 0.05 can obtain a slightly better result.

\noindent\textbf{Effect of Batch Size.} Fig.~\ref{fig:ablation}\red{D} explores the effect of batch size for training EMO. Small batch size ($\leq$ 512) will bring performance degradation, while high batch size will suffer from performance saturation, and it will also put higher requirements on the hardware. Therefore, 1,024 or 2,048 is enough to meet the training requirement.

% \subsection{Explanatory Analysis} \label{section:exp_analysis}

\begin{figure}[htp]
    \centering
    \includegraphics[width=1.0\linewidth]{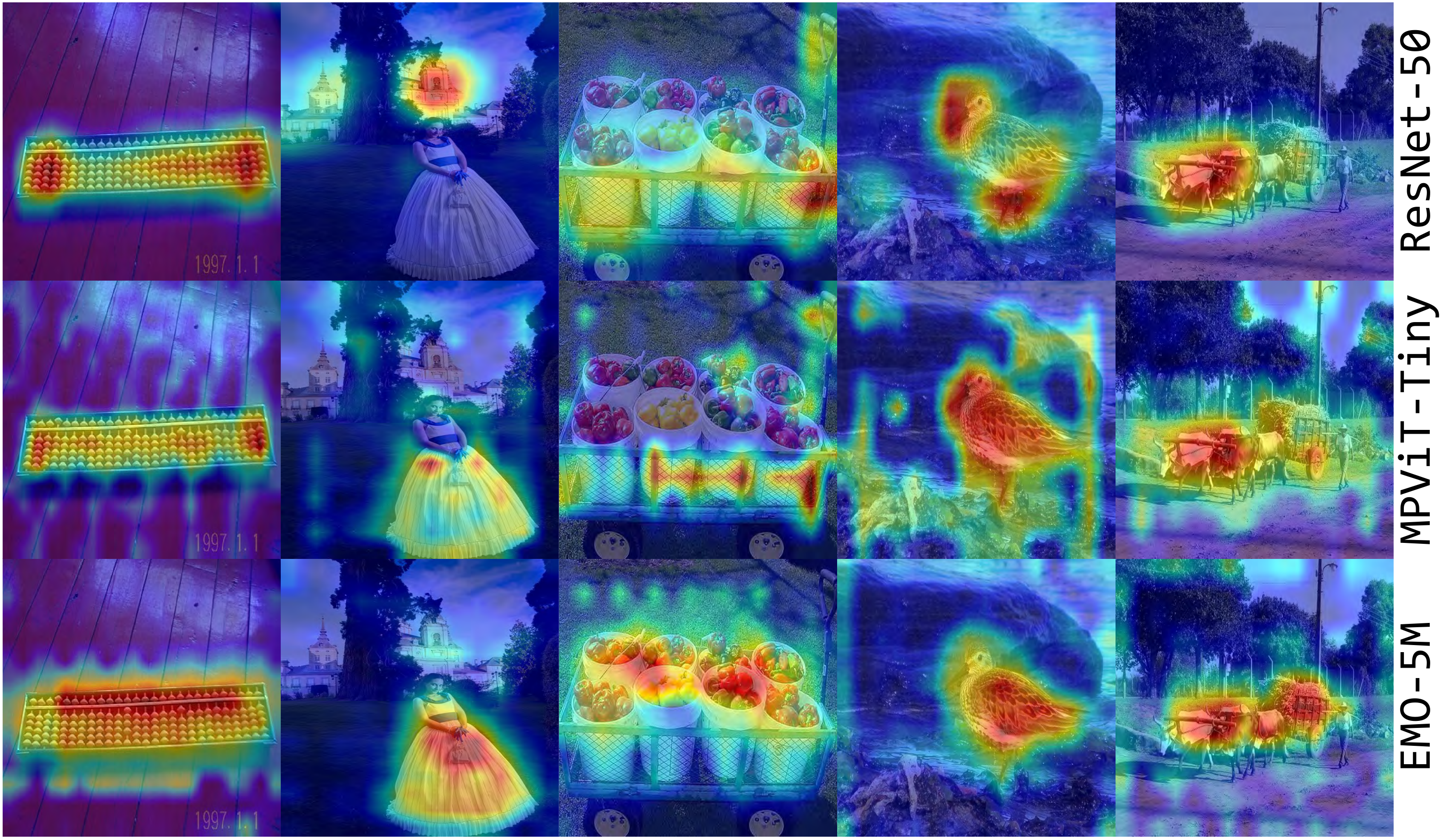}
    \caption{Visualizations by Grad-CAM among CNN-based ResNet, Transformer-based MPViT, and our EMO.}
    \label{fig:cam}
    \vspace{-1.0em}
\end{figure}

\noindent\textbf{Attention Visualizations by Grad-CAM.} To better illustrate the effectiveness of our approach, Grad-CAM~\cite{gradcam} is used to highlight concerning regions of different models. As shown in Fig.~\ref{fig:cam}, CNN-based ResNet tends to focus on specific objects, and Transformer-based MPViT pays more attention to global features. Comparatively, our EMO could focus more accurately on salient objects while keeping the capability of perceiving global regions. This potentially explains why EMO gets better results in various tasks.

\begin{wrapfigure}{r}{5.0cm}
    \centering
    \vspace{-0.8em}
    \includegraphics[width=1.0\linewidth]{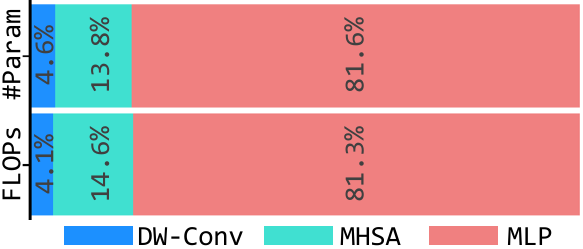}
    \caption{\#Params and FLOPs distributions of EMO-5M in terms of core modules in iRMB. MLP represents the expansion/shrinkage operations outside DW-Conv and MHSA.}
    \vspace{-1.0em}
    \label{fig:dist}
\end{wrapfigure}

\noindent\textbf{Distributions of \#Params and FLOPs.} iRMB mainly consists of DW-Conv and EW-MHSA modules, and Fig.~\ref{fig:dist} further displays distributions of \#Params and FLOPs. In general, DW-Conv and MHSA account for a low proportion of \#Params and FLOPs, \ie, 4.6\%/4.1\% and 13.8\%/14.6\%, respectively. Also, we found that \#Params is consistent with the proportion of FLOPs for our method, meaning that EMO is a relatively balanced model.

\begin{figure}[tp]
    \centering
    \includegraphics[width=1.0\linewidth]{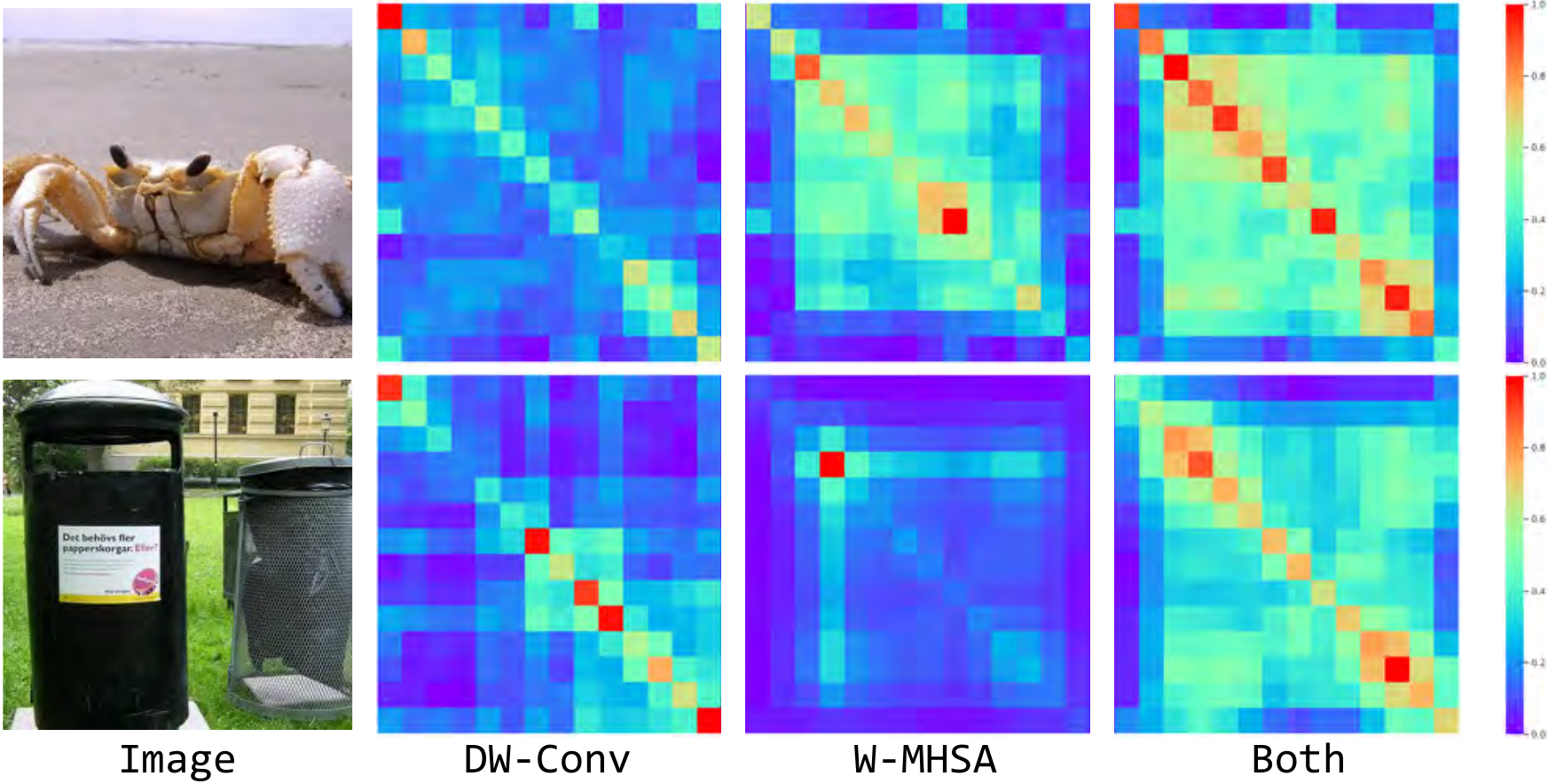}
    \caption{Diagonal similarity with different components.}
    \label{fig:irmb_attn}
    \vspace{-1.0em}
\end{figure}

\noindent\textbf{Feature Similarity Visualizations.} As mentioned in Sec.~\ref{section:irmb}, cascaded \emph{Convolution} and \emph{MHSA} operations can increase the expansion speed of the receptive field. To verify the validation of this design, we visualize the similarity of diagonal pixels in Stage-3 with different compositions, \ie, only DW-Conv, only EW-MHSA, and both modules. As shown in Fig.~\ref{fig:irmb_attn}, results show that features tend to have short-distance correlations when only DW-Conv is used, while EW-MHSA brings more long-distance correlations. Comparatively, iRMB takes advantage of both modules with a larger receptive field, \ie, distant locations have high similarities.

\section{Related Work} \label{section:related}
\noindent\textbf{Efficient CNN Models.}
With increasing demands of neural networks for mobile vision applications, efficient model designing has attracted extensive attention from researchers in recent years. SqueezeNet~\cite{squeezenet} replaces 3x3 filters with 1x1 filters and decreases channel numbers to reduce model parameters, while Inceptionv3~\cite{inceptionv3} factorizes the standard convolution into asymmetric convolutions. Later, MobileNet~\cite{mnetv1} introduces depth-wise separable convolution to alleviate a large amount of computation and parameters, followed in subsequent lightweight models~\cite{senet,mnetv2,shufflenetv1,shufflenetv2,espnetv2,ghostnet,sfnet,Li2022SFNetFA}. Besides the above hand-craft methods, researchers exploit automatic architecture design in the pre-defined search space~\cite{mnetv3,efficientnet,mnasnet,darts,ofa}. \\
\noindent\textbf{Hugging Vision Transformer with CNN.} Since ViT~\cite{vit} first introduces Transformer structure~\cite{transformer} into visual tasks, massive improvements have successfully been developed. DeiT~\cite{deit} provides a benchmark for efficient transformer training, subsequent works~\cite{pvtv1,pvtv2,swinv1} employ ResNet-like~\cite{resnet} pyramid structure to form pure Transformer-based models for dense prediction tasks. However, the absence of 2D convolution will potentially increase the optimization difficulty and damage the model accuracy for lacking local inductive bias, so researchers~\cite{hassanin2022visual,islam2022recent} concentrate on how to better integrate convolution into Transformer for obtaining stronger hybrid models. \Eg, work~\cite{ceit} incorporate convolution design into FFN, works~\cite{cpvt,uniformer} regard convolution as the positional embedding for enhancing inductive bias of the model, and works~\cite{coat,cvt} for attention and QKV calculations, respectively. Unlike the above methods that improve naive Transformer to obtain high performance, we study how to build a simple but effective lightweight model based on an improved one-residual attention block. \\
\noindent\textbf{Efficient Transformer Improvements.}
Recently, researchers have started to lighten Transformer-based models for low computational power. Tao~\etal~\cite{lightvit} introduce additional learnable tokens to capture global dependencies efficiently, and Chen~\etal~\cite{lightvit} design a parallel structure of MobileNet and Transformer with a two-way bridge in between. Works~\cite{rest,edgevit} improve an efficient Transformer block by borrowing convolution operation, while EdgeNeXt~\cite{edgenext} absorbs effective Res2Net~\cite{res2net} and transposed channel attention~\cite{xcit}. The recently popular MobileVit series~\cite{mvitv1,mvitv2,mvitv3} fuse improved MobileViT blocks with Mobile blocks~\cite{mnetv2} and achieve significant improvements over MobileNet~\cite{mnetv1,mnetv2,mnetv3} on several vision tasks. However, most current approaches build on transformer structure and require \textit{elaborate complex modules}, which limits the mobility and usability of the model. In summary, how to balance parameters, computation, and accuracy while designing an easy-to-use mobile model still needs further research.

\section{Conclusion and Future Works} \label{section:con}
This work rethinks lightweight infrastructure from efficient IRB and effective components of Transformer in a unified perspective, and we propose the concept of Meta Mobile Block for designing efficient models. In detail, we deduce a modern infrastructural iRMB and build a lightweight attention-based EMO with only iRMB for downstream tasks. Massive experiments on several datasets demonstrate the superiority of our approach. Also, we provide detailed studies of our method and give some experimental findings on building an attention-based lightweight model. Hope our study will inspire researchers to design more power efficient models and make interesting applications.

More complex operators may potentially improve the effectiveness of the model, \eg, transposed channel attention~\cite{xcit}, multi-scale Res2Net~\cite{res2net}, and efficient Performer~\cite{performer}, \etc, which should be thoroughly tried and experimented further to explore the upper limits of the efficient model structure. Also, higher resolution input, combined with Neural Architecture Search (NAS), distillation from heavy models, training on larger ImageNet-21K dataset, and stronger training augmentations/strategies~\cite{mvitv1,edgenext,token_labeling} will further improve the model performance. Limited by the current computational power, we will leave the above-mentioned attempts in our future works.

{\small
\bibliographystyle{ieee_fullname}
\bibliography{main}

\begin{thebibliography}{10}\itemsep=-1pt

\bibitem{xcit}
Alaaeldin Ali, Hugo Touvron, Mathilde Caron, Piotr Bojanowski, Matthijs Douze,
  Armand Joulin, Ivan Laptev, Natalia Neverova, Gabriel Synnaeve, Jakob
  Verbeek, et~al.
\newblock Xcit: Cross-covariance image transformers.
\newblock In {\em NeurIPS}, volume~34, 2021.

\bibitem{ln}
Jimmy~Lei Ba, Jamie~Ryan Kiros, and Geoffrey~E Hinton.
\newblock Layer normalization.
\newblock {\em arXiv preprint arXiv:1607.06450}, 2016.

\bibitem{ofa}
Han Cai, Chuang Gan, Tianzhe Wang, Zhekai Zhang, and Song Han.
\newblock Once-for-all: Train one network and specialize it for efficient
  deployment.
\newblock In {\em ICLR}, 2020.

\bibitem{mmdetection}
Kai Chen, Jiaqi Wang, Jiangmiao Pang, Yuhang Cao, Yu Xiong, Xiaoxiao Li,
  Shuyang Sun, Wansen Feng, Ziwei Liu, Jiarui Xu, Zheng Zhang, Dazhi Cheng,
  Chenchen Zhu, Tianheng Cheng, Qijie Zhao, Buyu Li, Xin Lu, Rui Zhu, Yue Wu,
  Jifeng Dai, Jingdong Wang, Jianping Shi, Wanli Ouyang, Chen~Change Loy, and
  Dahua Lin.
\newblock {MMDetection}: Open mmlab detection toolbox and benchmark.
\newblock {\em arXiv preprint arXiv:1906.07155}, 2019.

\bibitem{deeplabv3}
Liang-Chieh Chen, George Papandreou, Florian Schroff, and Hartwig Adam.
\newblock Rethinking atrous convolution for semantic image segmentation.
\newblock {\em arXiv preprint arXiv:1706.05587}, 2017.

\bibitem{mobileformer}
Yinpeng Chen, Xiyang Dai, Dongdong Chen, Mengchen Liu, Xiaoyi Dong, Lu Yuan,
  and Zicheng Liu.
\newblock Mobile-former: Bridging mobilenet and transformer.
\newblock In {\em CVPR}, 2022.

\bibitem{performer}
Krzysztof~Marcin Choromanski, Valerii Likhosherstov, David Dohan, Xingyou Song,
  Andreea Gane, Tamas Sarlos, Peter Hawkins, Jared~Quincy Davis, Afroz
  Mohiuddin, Lukasz Kaiser, David~Benjamin Belanger, Lucy~J Colwell, and Adrian
  Weller.
\newblock Rethinking attention with performers.
\newblock In {\em ICLR}, 2021.

\bibitem{cpvt}
Xiangxiang Chu, Zhi Tian, Bo Zhang, Xinlong Wang, and Chunhua Shen.
\newblock Conditional positional encodings for vision transformers.
\newblock In {\em ICLR}, 2023.

\bibitem{mmseg2020}
MMSegmentation Contributors.
\newblock {MMSegmentation}: Openmmlab semantic segmentation toolbox and
  benchmark.
\newblock \url{https://github.com/open-mmlab/mmsegmentation}, 2020.

\bibitem{randaugment}
Ekin~D Cubuk, Barret Zoph, Jonathon Shlens, and Quoc~V Le.
\newblock Randaugment: Practical automated data augmentation with a reduced
  search space.
\newblock In {\em CVPRW}, 2020.

\bibitem{imagenet}
Jia Deng, Wei Dong, Richard Socher, Li-Jia Li, Kai Li, and Li Fei-Fei.
\newblock Imagenet: A large-scale hierarchical image database.
\newblock In {\em CVPR}, 2009.

\bibitem{cswin}
Xiaoyi Dong, Jianmin Bao, Dongdong Chen, Weiming Zhang, Nenghai Yu, Lu Yuan,
  Dong Chen, and Baining Guo.
\newblock Cswin transformer: A general vision transformer backbone with
  cross-shaped windows.
\newblock In {\em CVPR}, 2022.

\bibitem{vit}
Alexey Dosovitskiy, Lucas Beyer, Alexander Kolesnikov, Dirk Weissenborn,
  Xiaohua Zhai, Thomas Unterthiner, Mostafa Dehghani, Matthias Minderer, Georg
  Heigold, Sylvain Gelly, Jakob Uszkoreit, and Neil Houlsby.
\newblock An image is worth 16x16 words: Transformers for image recognition at
  scale.
\newblock In {\em ICLR}, 2021.

\bibitem{res2net}
Shang-Hua Gao, Ming-Ming Cheng, Kai Zhao, Xin-Yu Zhang, Ming-Hsuan Yang, and
  Philip Torr.
\newblock Res2net: A new multi-scale backbone architecture.
\newblock {\em IEEE TPAMI}, 2019.

\bibitem{ghostnet}
Kai Han, Yunhe Wang, Qi Tian, Jianyuan Guo, Chunjing Xu, and Chang Xu.
\newblock Ghostnet: More features from cheap operations.
\newblock In {\em CVPR}, 2020.

\bibitem{hassanin2022visual}
Mohammed Hassanin, Saeed Anwar, Ibrahim Radwan, Fahad~S Khan, and Ajmal Mian.
\newblock Visual attention methods in deep learning: An in-depth survey.
\newblock {\em arXiv preprint arXiv:2204.07756}, 2022.

\bibitem{resnet}
Kaiming He, Xiangyu Zhang, Shaoqing Ren, and Jian Sun.
\newblock Deep residual learning for image recognition.
\newblock In {\em CVPR}, 2016.

\bibitem{gelu}
Dan Hendrycks and Kevin Gimpel.
\newblock Gaussian error linear units (gelus).
\newblock {\em arXiv preprint arXiv:1606.08415}, 2016.

\bibitem{mnetv3}
Andrew Howard, Mark Sandler, Grace Chu, Liang-Chieh Chen, Bo Chen, Mingxing
  Tan, Weijun Wang, Yukun Zhu, Ruoming Pang, Vijay Vasudevan, et~al.
\newblock Searching for mobilenetv3.
\newblock In {\em ICCV}, 2019.

\bibitem{mnetv1}
Andrew~G Howard, Menglong Zhu, Bo Chen, Dmitry Kalenichenko, Weijun Wang,
  Tobias Weyand, Marco Andreetto, and Hartwig Adam.
\newblock Mobilenets: Efficient convolutional neural networks for mobile vision
  applications.
\newblock {\em arXiv preprint arXiv:1704.04861}, 2017.

\bibitem{senet}
Jie Hu, Li Shen, and Gang Sun.
\newblock Squeeze-and-excitation networks.
\newblock In {\em CVPR}, 2018.

\bibitem{drop_path}
Gao Huang, Yu Sun, Zhuang Liu, Daniel Sedra, and Kilian~Q Weinberger.
\newblock Deep networks with stochastic depth.
\newblock In {\em ECCV}, 2016.

\bibitem{lightvit}
Tao Huang, Lang Huang, Shan You, Fei Wang, Chen Qian, and Chang Xu.
\newblock Lightvit: Towards light-weight convolution-free vision transformers.
\newblock {\em arXiv preprint arXiv:2207.05557}, 2022.

\bibitem{squeezenet}
Forrest~N Iandola, Song Han, Matthew~W Moskewicz, Khalid Ashraf, William~J
  Dally, and Kurt Keutzer.
\newblock Squeezenet: Alexnet-level accuracy with 50x fewer parameters and< 0.5
  mb model size.
\newblock {\em arXiv preprint arXiv:1602.07360}, 2016.

\bibitem{ios_cls}
Apple Inc.
\newblock Classifying images with vision and core ml.
\newblock
  \url{https://developer.apple.com/documentation/vision/classifying_images_with_vision_and_core_ml},
  2023.

\bibitem{bn}
Sergey Ioffe and Christian Szegedy.
\newblock Batch normalization: Accelerating deep network training by reducing
  internal covariate shift.
\newblock In {\em ICML}. PMLR, 2015.

\bibitem{islam2022recent}
Khawar Islam.
\newblock Recent advances in vision transformer: A survey and outlook of recent
  work.
\newblock {\em arXiv preprint arXiv:2203.01536}, 2022.

\bibitem{token_labeling}
Zi-Hang Jiang, Qibin Hou, Li Yuan, Daquan Zhou, Yujun Shi, Xiaojie Jin, Anran
  Wang, and Jiashi Feng.
\newblock All tokens matter: Token labeling for training better vision
  transformers.
\newblock In {\em NeurIPS}, volume~34, 2021.

\bibitem{reformer}
Nikita Kitaev, Lukasz Kaiser, and Anselm Levskaya.
\newblock Reformer: The efficient transformer.
\newblock In {\em ICLR}, 2020.

\bibitem{mpvit}
Youngwan Lee, Jonghee Kim, Jeffrey Willette, and Sung~Ju Hwang.
\newblock Mpvit: Multi-path vision transformer for dense prediction.
\newblock In {\em CVPR}, 2022.

\bibitem{nextvit}
Jiashi Li, Xin Xia, Wei Li, Huixia Li, Xing Wang, Xuefeng Xiao, Rui Wang, Min
  Zheng, and Xin Pan.
\newblock Next-vit: Next generation vision transformer for efficient deployment
  in realistic industrial scenarios.
\newblock {\em arXiv preprint arXiv:2207.05501}, 2022.

\bibitem{uniformer}
Kunchang Li, Yali Wang, Gao Peng, Guanglu Song, Yu Liu, Hongsheng Li, and Yu
  Qiao.
\newblock Uniformer: Unified transformer for efficient spatial-temporal
  representation learning.
\newblock In {\em ICLR}, 2022.

\bibitem{li2023transformer}
Xiangtai Li, Henghui Ding, Wenwei Zhang, Haobo Yuan, Guangliang Cheng, Pang
  Jiangmiao, Kai Chen, Ziwei Liu, and Chen~Change Loy.
\newblock Transformer-based visual segmentation: A survey.
\newblock {\em arXiv pre-print}, 2023.

\bibitem{sfnet}
Xiangtai Li, Ansheng You, Zhen Zhu, Houlong Zhao, Maoke Yang, Kuiyuan Yang, and
  Yunhai Tong.
\newblock Semantic flow for fast and accurate scene parsing.
\newblock In {\em ECCV}, 2020.

\bibitem{Li2022SFNetFA}
Xiangtai Li, Jiangning Zhang, Yibo Yang, Guangliang Cheng, Kuiyuan Yang, Yu
  Tong, and Dacheng Tao.
\newblock Sfnet: Faster, accurate, and domain agnostic semantic segmentation
  via semantic flow.
\newblock {\em ArXiv}, 2022.

\bibitem{efficientformer}
Yanyu Li, Geng Yuan, Yang Wen, Ju Hu, Georgios Evangelidis, Sergey Tulyakov,
  Yanzhi Wang, and Jian Ren.
\newblock Efficientformer: Vision transformers at mobilenet speed.
\newblock {\em Advances in Neural Information Processing Systems},
  35:12934--12949, 2022.

\bibitem{retinanet}
Tsung-Yi Lin, Priya Goyal, Ross Girshick, Kaiming He, and Piotr Doll{\'a}r.
\newblock Focal loss for dense object detection.
\newblock In {\em ICCV}, 2017.

\bibitem{coco}
Tsung-Yi Lin, Michael Maire, Serge Belongie, James Hays, Pietro Perona, Deva
  Ramanan, Piotr Doll{\'a}r, and C~Lawrence Zitnick.
\newblock Microsoft coco: Common objects in context.
\newblock In {\em ECCV}, 2014.

\bibitem{gmlp}
Hanxiao Liu, Zihang Dai, David So, and Quoc~V Le.
\newblock Pay attention to mlps.
\newblock {\em NeurIPS}, 2021.

\bibitem{darts}
Hanxiao Liu, Karen Simonyan, and Yiming Yang.
\newblock {DARTS}: Differentiable architecture search.
\newblock In {\em ICLR}, 2019.

\bibitem{swinv2}
Ze Liu, Han Hu, Yutong Lin, Zhuliang Yao, Zhenda Xie, Yixuan Wei, Jia Ning, Yue
  Cao, Zheng Zhang, Li Dong, et~al.
\newblock Swin transformer v2: Scaling up capacity and resolution.
\newblock In {\em CVPR}, 2022.

\bibitem{swinv1}
Ze Liu, Yutong Lin, Yue Cao, Han Hu, Yixuan Wei, Zheng Zhang, Stephen Lin, and
  Baining Guo.
\newblock Swin transformer: Hierarchical vision transformer using shifted
  windows.
\newblock In {\em ICCV}, 2021.

\bibitem{cosine}
Ilya Loshchilov and Frank Hutter.
\newblock {SGDR}: Stochastic gradient descent with warm restarts.
\newblock In {\em ICLR}, 2017.

\bibitem{adamw}
Ilya Loshchilov and Frank Hutter.
\newblock Decoupled weight decay regularization.
\newblock In {\em ICLR}, 2019.

\bibitem{mocovit}
Hailong Ma, Xin Xia, Xing Wang, Xuefeng Xiao, Jiashi Li, and Min Zheng.
\newblock Mocovit: Mobile convolutional vision transformer.
\newblock {\em arXiv preprint arXiv:2205.12635}, 2022.

\bibitem{shufflenetv2}
Ningning Ma, Xiangyu Zhang, Hai-Tao Zheng, and Jian Sun.
\newblock Shufflenet v2: Practical guidelines for efficient cnn architecture
  design.
\newblock In {\em ECCV}, 2018.

\bibitem{edgenext}
Muhammad Maaz, Abdelrahman Shaker, Hisham Cholakkal, Salman Khan, Syed~Waqas
  Zamir, Rao~Muhammad Anwer, and Fahad~Shahbaz Khan.
\newblock Edgenext: efficiently amalgamated cnn-transformer architecture for
  mobile vision applications.
\newblock In {\em ECCVW}, 2022.

\bibitem{delight}
Sachin Mehta, Marjan Ghazvininejad, Srinivasan Iyer, Luke Zettlemoyer, and
  Hannaneh Hajishirzi.
\newblock Delight: Deep and light-weight transformer.
\newblock In {\em ICLR}, 2021.

\bibitem{mvitv1}
Sachin Mehta and Mohammad Rastegari.
\newblock Mobilevit: Light-weight, general-purpose, and mobile-friendly vision
  transformer.
\newblock In {\em ICLR}, 2022.

\bibitem{mvitv2}
Sachin Mehta and Mohammad Rastegari.
\newblock Separable self-attention for mobile vision transformers.
\newblock {\em arXiv preprint arXiv:2206.02680}, 2022.

\bibitem{espnetv2}
Sachin Mehta, Mohammad Rastegari, Linda Shapiro, and Hannaneh Hajishirzi.
\newblock Espnetv2: A light-weight, power efficient, and general purpose
  convolutional neural network.
\newblock In {\em CVPR}, 2019.

\bibitem{edgevit}
Junting Pan, Adrian Bulat, Fuwen Tan, Xiatian Zhu, Lukasz Dudziak, Hongsheng
  Li, Georgios Tzimiropoulos, and Brais Martinez.
\newblock Edgevits: Competing light-weight cnns on mobile devices with vision
  transformers.
\newblock In {\em ECCV}, 2022.

\bibitem{pytorch}
Adam Paszke, Sam Gross, Francisco Massa, Adam Lerer, James Bradbury, Gregory
  Chanan, Trevor Killeen, Zeming Lin, Natalia Gimelshein, Luca Antiga, et~al.
\newblock Pytorch: An imperative style, high-performance deep learning library.
\newblock In {\em NeurIPS}, volume~32, 2019.

\bibitem{mnetv2}
Mark Sandler, Andrew Howard, Menglong Zhu, Andrey Zhmoginov, and Liang-Chieh
  Chen.
\newblock Mobilenetv2: Inverted residuals and linear bottlenecks.
\newblock In {\em CVPR}, 2018.

\bibitem{gradcam}
Ramprasaath~R Selvaraju, Michael Cogswell, Abhishek Das, Ramakrishna Vedantam,
  Devi Parikh, and Dhruv Batra.
\newblock Grad-cam: Visual explanations from deep networks via gradient-based
  localization.
\newblock In {\em ICCV}, 2017.

\bibitem{inception}
Chenyang Si, Weihao Yu, Pan Zhou, Yichen Zhou, Xinchao Wang, and Shuicheng YAN.
\newblock Inception transformer.
\newblock In {\em NeurIPS}, 2022.

\bibitem{dropout}
Nitish Srivastava, Geoffrey Hinton, Alex Krizhevsky, Ilya Sutskever, and Ruslan
  Salakhutdinov.
\newblock Dropout: a simple way to prevent neural networks from overfitting.
\newblock 2014.

\bibitem{inceptionv3}
Christian Szegedy, Vincent Vanhoucke, Sergey Ioffe, Jon Shlens, and Zbigniew
  Wojna.
\newblock Rethinking the inception architecture for computer vision.
\newblock In {\em CVPR}, 2016.

\bibitem{label_smoothing}
Christian Szegedy, Vincent Vanhoucke, Sergey Ioffe, Jon Shlens, and Zbigniew
  Wojna.
\newblock Rethinking the inception architecture for computer vision.
\newblock In {\em CVPR}, 2016.

\bibitem{mnasnet}
Mingxing Tan, Bo Chen, Ruoming Pang, Vijay Vasudevan, Mark Sandler, Andrew
  Howard, and Quoc~V Le.
\newblock Mnasnet: Platform-aware neural architecture search for mobile.
\newblock In {\em CVPR}, 2019.

\bibitem{efficientnet}
Mingxing Tan and Quoc Le.
\newblock Efficientnet: Rethinking model scaling for convolutional neural
  networks.
\newblock In {\em ICML}. PMLR, 2019.

\bibitem{mlpmixer}
Ilya~O Tolstikhin, Neil Houlsby, Alexander Kolesnikov, Lucas Beyer, Xiaohua
  Zhai, Thomas Unterthiner, Jessica Yung, Andreas Steiner, Daniel Keysers,
  Jakob Uszkoreit, et~al.
\newblock Mlp-mixer: An all-mlp architecture for vision.
\newblock {\em NeurIPS}, 2021.

\bibitem{resmlp}
Hugo Touvron, Piotr Bojanowski, Mathilde Caron, Matthieu Cord, Alaaeldin
  El-Nouby, Edouard Grave, Gautier Izacard, Armand Joulin, Gabriel Synnaeve,
  Jakob Verbeek, et~al.
\newblock Resmlp: Feedforward networks for image classification with
  data-efficient training.
\newblock {\em IEEE Transactions on Pattern Analysis and Machine Intelligence},
  2022.

\bibitem{deit}
Hugo Touvron, Matthieu Cord, Matthijs Douze, Francisco Massa, Alexandre
  Sablayrolles, and Herv{\'e} J{\'e}gou.
\newblock Training data-efficient image transformers \& distillation through
  attention.
\newblock In {\em ICML}, 2021.

\bibitem{ls}
Hugo Touvron, Matthieu Cord, Alexandre Sablayrolles, Gabriel Synnaeve, and
  Herv{\'e} J{\'e}gou.
\newblock Going deeper with image transformers.
\newblock In {\em ICCV}, 2021.

\bibitem{transformer}
Ashish Vaswani, Noam Shazeer, Niki Parmar, Jakob Uszkoreit, Llion Jones,
  Aidan~N Gomez, {\L}ukasz Kaiser, and Illia Polosukhin.
\newblock Attention is all you need.
\newblock 30, 2017.

\bibitem{mvitv3}
Shakti~N Wadekar and Abhishek Chaurasia.
\newblock Mobilevitv3: Mobile-friendly vision transformer with simple and
  effective fusion of local, global and input features.
\newblock {\em arXiv preprint arXiv:2209.15159}, 2022.

\bibitem{pvtv1}
Wenhai Wang, Enze Xie, Xiang Li, Deng-Ping Fan, Kaitao Song, Ding Liang, Tong
  Lu, Ping Luo, and Ling Shao.
\newblock Pyramid vision transformer: A versatile backbone for dense prediction
  without convolutions.
\newblock In {\em ICCV}, 2021.

\bibitem{pvtv2}
Wenhai Wang, Enze Xie, Xiang Li, Deng-Ping Fan, Kaitao Song, Ding Liang, Tong
  Lu, Ping Luo, and Ling Shao.
\newblock Pvt v2: Improved baselines with pyramid vision transformer.
\newblock {\em CVM}, 2022.

\bibitem{timm}
Ross Wightman.
\newblock Pytorch image models.
\newblock \url{https://github.com/rwightman/pytorch-image-models}, 2019.

\bibitem{cvt}
Haiping Wu, Bin Xiao, Noel Codella, Mengchen Liu, Xiyang Dai, Lu Yuan, and Lei
  Zhang.
\newblock Cvt: Introducing convolutions to vision transformers.
\newblock In {\em ICCV}, 2021.

\bibitem{coat}
Weijian Xu, Yifan Xu, Tyler Chang, and Zhuowen Tu.
\newblock Co-scale conv-attentional image transformers.
\newblock In {\em ICCV}, 2021.

\bibitem{moat}
Chenglin Yang, Siyuan Qiao, Qihang Yu, Xiaoding Yuan, Yukun Zhu, Alan Yuille,
  Hartwig Adam, and Liang-Chieh Chen.
\newblock Moat: Alternating mobile convolution and attention brings strong
  vision models.
\newblock {\em ICLR}, 2023.

\bibitem{focal}
Jianwei Yang, Chunyuan Li, Pengchuan Zhang, Xiyang Dai, Bin Xiao, Lu Yuan, and
  Jianfeng Gao.
\newblock Focal attention for long-range interactions in vision transformers.
\newblock In {\em NeurIPS}, 2021.

\bibitem{metaformer}
Weihao Yu, Mi Luo, Pan Zhou, Chenyang Si, Yichen Zhou, Xinchao Wang, Jiashi
  Feng, and Shuicheng Yan.
\newblock Metaformer is actually what you need for vision.
\newblock In {\em CVPR}, 2022.

\bibitem{ceit}
Kun Yuan, Shaopeng Guo, Ziwei Liu, Aojun Zhou, Fengwei Yu, and Wei Wu.
\newblock Incorporating convolution designs into visual transformers.
\newblock In {\em ICCV}, 2021.

\bibitem{cutmix}
Sangdoo Yun, Dongyoon Han, Seong~Joon Oh, Sanghyuk Chun, Junsuk Choe, and
  Youngjoon Yoo.
\newblock Cutmix: Regularization strategy to train strong classifiers with
  localizable features.
\newblock In {\em ICCV}, 2019.

\bibitem{mixup}
Hongyi Zhang, Moustapha Cisse, Yann~N. Dauphin, and David Lopez-Paz.
\newblock mixup: Beyond empirical risk minimization.
\newblock In {\em ICLR}, 2018.

\bibitem{zhang2022eatformer}
Jiangning Zhang, Xiangtai Li, Yabiao Wang, Chengjie Wang, Yibo Yang, Yong Liu,
  and Dacheng Tao.
\newblock Eatformer: improving vision transformer inspired by evolutionary
  algorithm.
\newblock {\em arXiv preprint arXiv:2206.09325}, 2022.

\bibitem{zhang2021analogous}
Jiangning Zhang, Chao Xu, Jian Li, Wenzhou Chen, Yabiao Wang, Ying Tai, Shuo
  Chen, Chengjie Wang, Feiyue Huang, and Yong Liu.
\newblock Analogous to evolutionary algorithm: Designing a unified sequence
  model.
\newblock {\em NeurIPS}, 2021.

\bibitem{rest}
Qinglong Zhang and Yu-Bin Yang.
\newblock Rest: An efficient transformer for visual recognition.
\newblock 2021.

\bibitem{shufflenetv1}
Xiangyu Zhang, Xinyu Zhou, Mengxiao Lin, and Jian Sun.
\newblock Shufflenet: An extremely efficient convolutional neural network for
  mobile devices.
\newblock In {\em CVPR}, 2018.

\bibitem{pspnet}
Hengshuang Zhao, Jianping Shi, Xiaojuan Qi, Xiaogang Wang, and Jiaya Jia.
\newblock Pyramid scene parsing network.
\newblock In {\em CVPR}, 2017.

\bibitem{random_erasing}
Zhun Zhong, Liang Zheng, Guoliang Kang, Shaozi Li, and Yi Yang.
\newblock Random erasing data augmentation.
\newblock In {\em AAAI}, 2020.

\bibitem{ade20k}
Bolei Zhou, Hang Zhao, Xavier Puig, Tete Xiao, Sanja Fidler, Adela Barriuso,
  and Antonio Torralba.
\newblock Semantic understanding of scenes through the ade20k dataset.
\newblock {\em IJCV}, 2019.

\end{thebibliography}
}
\end{document}